\documentclass[pdflatex,sn-mathphys-num,iicol]{sn-jnl}

\usepackage{graphicx}%
\usepackage{enumitem}%
\usepackage{multirow}%
\usepackage{amsmath,amssymb,amsfonts}%
\usepackage{amsthm}%
\usepackage{mathrsfs}%
\usepackage[title]{appendix}%
\usepackage{xcolor}%
\usepackage{textcomp}%
\usepackage{manyfoot}%
\usepackage{booktabs}%
\usepackage{algorithm}%
\usepackage{algorithmicx}%
\usepackage{algpseudocode}%
\usepackage{listings}%
\usepackage{multirow}
\usepackage[table]{xcolor}
\usepackage{subcaption}
\usepackage{float}

\usepackage[font=footnotesize,labelfont=footnotesize]{caption}
\usepackage[font=footnotesize,labelfont=footnotesize]{subcaption}

\theoremstyle{thmstyleone}%
\theoremstyle{thmstyletwo}%

\theoremstyle{thmstylethree}%

\begin{document}

\title[FORGE]{FORGE: Frame Orthogonality in Relevance Geometry for Long-Form Video Understanding}

\author*[1]{\fnm{Ghazal} \sur{Kaviani}}\email{gkaviani3@gatech.edu}

\author[1]{\fnm{Ghassan} \sur{AlRegib}} 

\affil[1]{%
  \orgdiv{OLIVES at the Center for Signal and Information Processing (CSIP), School of Electrical and Computer Engineering},
  \orgname{Georgia Institute of Technology},
  \orgaddress{%
    \city{Atlanta},
    \state{GA},
    \country{USA}}%
}

\abstract{Multimodal large language models (MLLMs) have enabled long-form video understanding at a scale that was not previously possible. However, the density of relevant content decreases sharply as video sequence length increases, and exposing the model to more irrelevant content measurably reduces its accuracy. In this paper, we address the problem of maximizing query-relevant information in a frame subset selected at inference time, without training. \textbf{FORGE} (\textit{Frame Orthogonality in Relevance Geometry}) is a \emph{model-agnostic} method that induces a query-conditioned geometry on a pretrained multimodal embedding space, unifying relevance and diversity into a single objective. In this space, frames that cover independent query-relevant directions are far apart, and selecting the subset of maximum information captures diverse query-relevant content within the budget. Experiments on Video-MME and LongVideoBench at budgets of 16, 32, and 64 frames show that FORGE improves the unified keyframe selection score by 11.0–15.3 points over the strongest training-free baseline and up to doubles keyframe recall (0.415 vs. 0.204 at K=64 on Video-MME). The gains extend to question answering, where accuracy improves in every evaluated setting across eight open-source MLLMs spanning 4B to 32B parameters, by up to 8.7 points over uniform sampling and 5.2 points over the strongest baseline. Our findings suggest that aligning the embedding space with the query's high-dimensional structure is a promising direction for inference-time video understanding.}

\keywords{long-form video understanding, video question answering, keyframe selection, mutual information, vision-language models, multimodal large language models}

\onecolumn
\thispagestyle{empty}

\begin{description}[labelindent=0cm,leftmargin=3.3cm,labelwidth=3.1cm,style=multiline,itemsep=0.5em]

\item[\textbf{Citation}]{G. Kaviani and G. AlRegib, ``FORGE: Frame Orthogonality in
Relevance Geometry for Long-Form Video Understanding,'' arXiv:2607.25266 [cs.CV], 2026.}

\item[\textbf{DOI}]{\url{https://doi.org/10.48550/arXiv.2607.25266}}

\item[\textbf{Preprint}]{\url{https://arxiv.org/abs/2607.25266}}

\item[\textbf{Review}]{Under review at the \emph{International Journal of Computer Vision}.}

\item[\textbf{Code \& Data}]{\url{https://alregib.ece.gatech.edu/software-and-datasets/}}

\item[\textbf{Bib}]{@article\{kaviani2026forge,\\
title=\{FORGE: Frame Orthogonality in Relevance Geometry for Long-Form Video Understanding\},\\
author=\{Ghazal Kaviani and Ghassan AlRegib\},\\
year=\{2026\},\\
eprint=\{2607.25266\},\\
archivePrefix=\{arXiv\},\\
primaryClass=\{cs.CV\},\\
doi=\{10.48550/arXiv.2607.25266\},\\
url=\{https://arxiv.org/abs/2607.25266\},\}}

\item[\textbf{Copyright}]{\textcopyright{} 2026 The Authors. This is a preprint of a
manuscript currently under review. It is distributed under the arXiv.org perpetual,
non-exclusive license.}

\item[\textbf{Contact}]{\href{mailto:alregib@gatech.edu}{alregib@gatech.edu} \\
\url{https://alregib.ece.gatech.edu/}}

\end{description}

\newpage
\clearpage
\setcounter{page}{1}

\maketitle
\twocolumn

\section{Introduction}
\label{sec:intro}

\begin{figure*}[!htbp]
    \centering
    \includegraphics[width=0.95\textwidth]{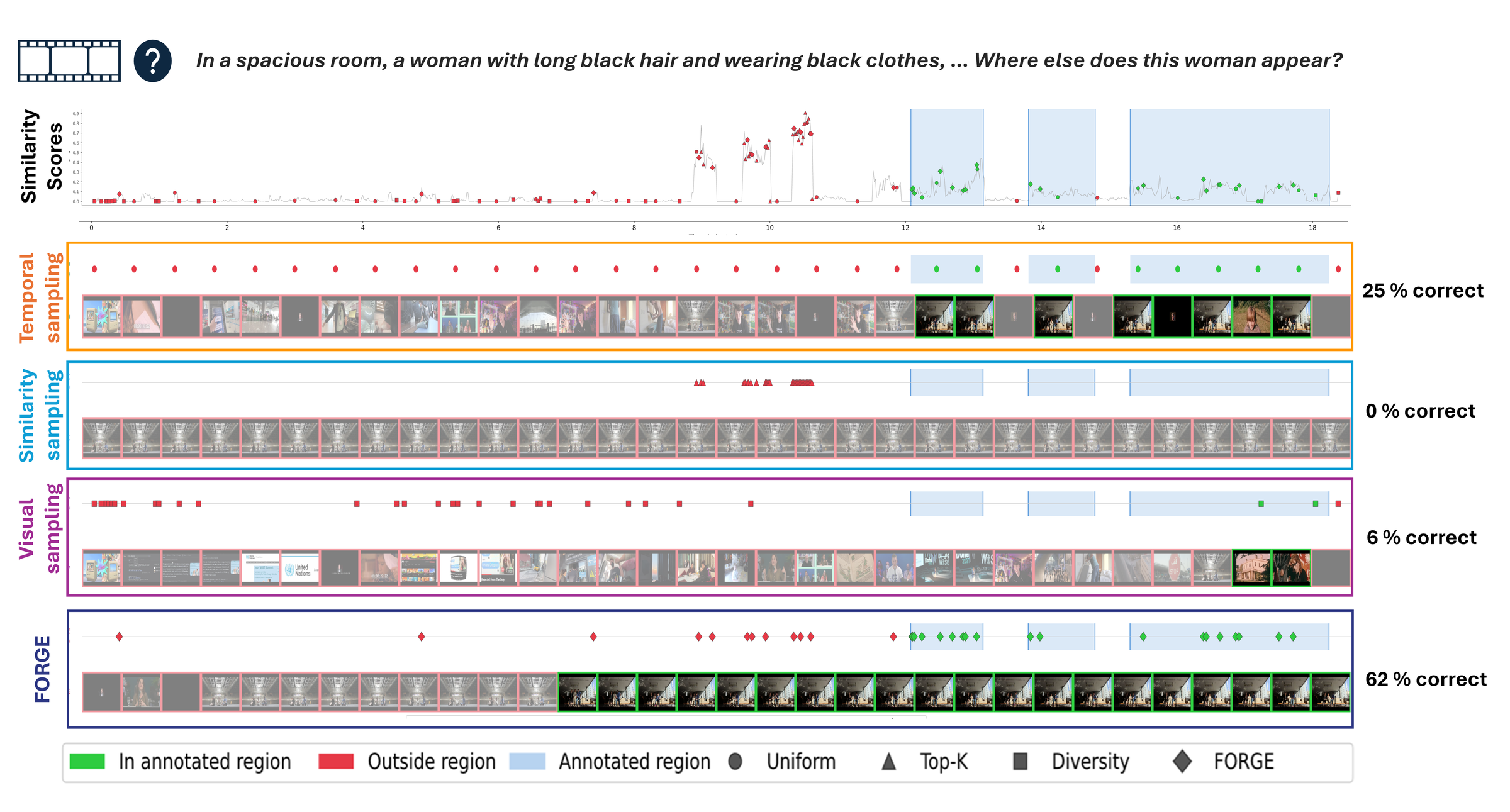}
    \caption{Scalar collapse in three training-free selectors, shown on a single query and video. The top row shows raw query-frame similarity along the timeline; high scores do not reliably coincide with the annotated answer region (shaded). Temporal sampling selects by position, similarity-based sampling by a scalar query-match score, and visual sampling by diversity in a query-agnostic embedding, recovering 25\%, 0\%, and 6\% of their frames from inside the annotated region respectively (red borders). \textbf{FORGE} induces a query-conditioned geometry that unifies the three signals into a single selection objective and recovers 62\% (green borders).}
    \label{fig:Fig_main}
\end{figure*}

Multimodal large language models (MLLMs) are currently the dominant approach to video understanding. Their recent advancements extend the application from short clips to long-form videos spanning several minutes to hours~\cite{lin2024video,maaz2024video,bai2025qwen2,chen2024far}. MLLMs face a persistent challenge with any long-form input; the difficulty is compounded in long-form video because task-relevant information becomes increasingly sparse as the sequence length grows. The evidence needed to answer a typical question is sparse, spanning a median of 2.1\% of the duration on Video-MME and 1.5\% on LongVideoBench~\cite{fu2025video,wu2024longvideobench,li2026kfs}. At inference time, the practical question is not how much of the video an MLLM can see, but which part of it contains the information the model needs.

Over the years, different approaches have been proposed to address long video understanding. Some build long-video MLLMs that extend the context window to process more frames at once~\cite{shu2025video,wang2024videollamb,jiang2025storm}. Processing more frames, however, does not guarantee better understanding when irrelevant content degrades attention to relevant regions~\cite{liu2024lost,du-etal-2025-context}. Another line of work compresses or prunes tokens to fit more content into the same budget~\cite{Yang2025CVPR,shen2025fastvid,adaretake}. Token compression is largely orthogonal to the choice of input frames and can be applied on top of most other approaches, including ours. A third line uses agentic and LLM-in-the-loop pipelines that make iterative or several separate calls to an MLLM over parts of the video to caption, summarize, or provide a judgment, and combine the results into a final answer~\cite{wang2024videoagent,zhang2025deep,zou2025air}. These pipelines perform well, but the repeated MLLM calls make them substantially costly~\cite{wang2025avp}, and scale poorly on consumer hardware. A fourth line is training-free frame selection: the MLLM stays fixed and a subset of frames is picked at inference time, replacing the uniform-sampling protocol that long-video MLLMs default to when no selector is applied~\cite{tang2025adaptive,Sun_2025_ICCV,liu2025bolt,li2025maxinfo,Zhang_2025_ICCV}. We work in this family. It requires no training, runs the MLLM once, and is model-agnostic, which makes it portable across MLLMs and compatible with the other approaches.

Given a fixed frame budget, the objective at inference time becomes selecting the maximum relevant information within that budget for a query, i.e., the natural language question the MLLM is asked to answer. Prior methods reduce each frame's query-relevance to a scalar and then spread their choices along a separate space, e.g., temporal position, visual features, or cluster membership. But the query embedding is a high-dimensional vector where frames can be close to one another along some dimensions and far along others. Collapsing the query's rank to a scalar, which we refer to as \emph{scalar collapse}, loses this directional structure and flattens the relevance from a full vector to a number. Consequently, the selection that follows cannot tell which query dimensions a frame covers and which it does not; frames concentrate on whichever dimensions dominate the scalar, while the evidence the answer requires often lies along the remaining dimensions and goes unselected. Because that evidence occupies only a narrow interval of the video, the loss is visible on the timeline. Figure~\ref{fig:Fig_main} illustrates scalar collapse in three selectors on a single query whose annotated answer region is shaded. Temporal sampling spreads picks along the timeline and recovers 25\% of its frames from inside the region, similarity-based sampling ranks frames by their pointwise match to the query and recovers none, and visual sampling diversifies frames in a query-agnostic embedding and recovers 6\%. None of the three reliably place frames inside the answer region, because a high similarity score indicates that a frame resembles the query rather than that it carries the evidence the answer requires.

We propose \textbf{FORGE} \textit{(Frame Orthogonality in Relevance Geometry)}, an inference-time, model-agnostic method that maximizes query-relevant information within a selection budget. \textbf{FORGE} builds a query-conditioned geometry in which frame orthogonality reflects independent query-relevant content, and selects $K$ frames by maximizing the information content of a subspace in this geometry. To our knowledge, \textbf{FORGE} is the first training-free frame selector that preserves the query's representational structure on the geometry where selection happens, so that relevance and diversity are unified into a single selection objective rather than computed as two objectives on separate feature spaces. On Video-MME~\cite{fu2025video} and LongVideoBench~\cite{wu2024longvideobench}, \textbf{FORGE} improves keyframe recall, scene coverage, and their unified selection score over five representative baselines in every evaluated set, and these improvements translate to higher downstream accuracy on video question answering across a comprehensive set of open-source MLLMs.

Our contributions are as follows:
\begin{itemize}
    \item We identify \emph{scalar collapse} in existing training-free selectors, which reduce query-relevance to a per-frame scalar balanced against a query-agnostic spread on another feature, e.g., temporal position, visual features, or cluster membership.
    \item We propose \textbf{FORGE}, a training-free, model-agnostic method that induces a query-conditioned geometry on the frame embeddings, unifying relevance and diversity into a single selection objective.
    \item \textbf{FORGE} consistently outperforms five representative baselines on relevance, coverage, and a unified score on Video-MME and LongVideoBench, and on downstream accuracy on long-video question answering across a comprehensive set of open-source MLLMs.
\end{itemize}

\section{Related Work}
\label{sec:related}


\subsection*{Long-video understanding in the MLLM era}
\label{sec:related-mllm}

Video understanding has progressed from clip-level recognition into the realm of multimodal large language models, in which a sequence of frame embeddings is processed with a text query and the output is produced based on both inputs. The datasets and benchmarks have expanded accordingly, with recent multi-task video question-answering benchmarks testing perception, temporal reasoning, grounding, and retrieval across video samples that range from short clips to hour-long videos. Research in this domain is shaped by a few coupled constraints: the length of the input the model is given, the amount of content the model can hold, and the reasoning the model is asked to perform, and the four paradigms reviewed below correspond to solutions targeted at these constraints.

\subsection*{Long-context and memory-based MLLMs}
\label{sec:related-longcontext}

Increasing the length of input that an MLLM can handle has motivated a family of architectural changes. Video-XL~\cite{shu2025video} and LongVLM~\cite{weng2024longvlm} extend the context window for hour-scale inputs, while VideoLLaMB~\cite{wang2024videollamb} introduces recurrent memory bridges that maintain long-term state across a streaming input. SeViLA~\cite{yu2023self} chains image-language models to localize and answer over longer videos. General-purpose MLLMs such as Qwen2.5-VL~\cite{bai2025qwen2}, Qwen3-VL~\cite{bai2025qwen3vl}, Video-LLaVA~\cite{lin2024video}, Video-ChatGPT~\cite{maaz2024video}, and InternVL~\cite{chen2024far} act as the downstream backbones on which most long-video methods are evaluated. \textbf{FORGE} is a model-agnostic method that does not modify the backbone, enlarge the context, or change the memory mechanism, but it can compose with any of the MLLMs above and more.

\subsection*{Token compression and pruning}
\label{sec:related-compression}

A complementary paradigm addresses the input length challenge at the level of the token stream. VisionZip~\cite{Yang2025CVPR} and VFlowOpt~\cite{Yang_2025_ICCV} prune tokens using information-flow signals inside the MLLM; FastVID~\cite{shen2025fastvid} applies dynamic density pruning; AdaReTaKe~\cite{adaretake} and PruneVid~\cite{prunevid} reduce temporal and cross-frame redundancy; and SparseVLM~\cite{zhang2025sparsevlm}, FlowCut~\cite{tong2025flowcut}, DynamicViT~\cite{rao2021dynamicvit}, and the spatial-temporal token selector of Wang et~al.~\cite{wang2022efficient} operate at the transformer-block level. These methods compress or prune the token stream the MLLM has already received, independent of which parts of the video carry the information the query requires, and they benefit from being fed a stream with a higher density of query-relevant content. \textbf{FORGE} is complementary to these approaches on both counts.

\subsection*{Agentic and iterative pipelines}
\label{sec:related-agentic}

Another approach is to treat video understanding as an interactive process in which an LLM or MLLM is repeatedly invoked as a reasoner over parts of the video. VideoAgent~\cite{wang2024videoagent} and its memory-augmented variant~\cite{fan2024videoagent} caption and re-query clips in a loop; Deep Video Discovery~\cite{zhang2025deep} and AIR~\cite{zou2025air} build adaptive tool-use and reasoning loops that decide what to read next; Active Video Perception~\cite{wang2025avp} iterates evidence seeking and reports cost reductions relative to exhaustive agentic baselines. Frame-Voyager~\cite{yu2024frame} and Re-thinking Temporal Search~\cite{ye2025re} cast selection itself as a learned or search-driven procedure, and VideoTree~\cite{wang2025videotree} organizes the video into a hierarchical representation that an LLM reasons over. \textbf{FORGE} does not invoke the MLLM during selection and does not iterate; it produces its subset in a single forward pass over the frame embeddings and the query, at a fraction of the compute these pipelines require (Section~\ref{sec:experiments}).

\subsection*{Frame selection}
\label{sec:related-selection}

Closest to the present work is the literature on frame selection itself, which splits cleanly into supervised and training-free categories.

\paragraph{Supervised and task-specific selectors.}
The supervised lineage trains a selector from task annotations. AdaFrame~\cite{wu2019adaframe} and MARL~\cite{wu2019MARL} learn sampling policies via reinforcement signals; ClipBERT~\cite{lei2021less} sparsely samples clips for end-to-end video-language training; Flexible Frame Selection~\cite{buch2025flexible} learns how many frames each video requires; M-LLM-based selection~\cite{Hu_2025_CVPR} fine-tunes an MLLM to choose frames for downstream question answering. Adjacent to this line, video summarization was not designed for long-video question answering but operates under the same input interface long-video MLLMs expect, producing a short subset from a long video that can be consumed by the MLLM without modification. Including a summarization baseline therefore serves as a direct test of whether a generic summary preserves the information the MLLM needs to answer the query, or discards too much evidence in the process. CSTA~\cite{son2024csta} and the spatiotemporal ViT of Hsu et~al.~\cite{10124837} train attention-based summarizers from labeled summaries, and query-focused video summarization addresses the query-aware variant of this problem: category-specific selection~\cite{10.1007/978-3-319-10599-4_35}, submodular mixtures over objectives~\cite{gygli2015video}, diverse sequential subset selection~\cite{gong2014diverse}, and query-focused or query-controllable summarization~\cite{sharghi2017query,huang2020query,huang2023query}. All of these methods require labeled summaries, saliency annotations, or task supervision that long-video MLLM benchmarks do not provide at inference time.

\paragraph{Training-free selectors.}
The training-free lineage operates with a frozen visual encoder and a frozen MLLM, deriving the subset of frames from the video and the query alone. We organize existing methods by the axis along which they spread their picks. A first group ranks frames by a query-grounded relevance scalar and takes the top-$K$ with no additional spreading: TiFRe~\cite{zheng2026tifre} parses the query into target objects and ranks frames by CLIP similarity to those objects, and Divide-then-Ground~\cite{li2025dividegroundadaptingframe} adapts the selection to the query type before ranking. A second group combines a relevance scalar with spreading along the temporal axis: AKS~\cite{tang2025adaptive} selects relevance peaks under a coverage constraint; BOLT~\cite{liu2025bolt} applies inverse-transform sampling to the relevance distribution; Think-Clip-Sample~\cite{tan2026think} allocates a slow-fast budget across clips under query guidance. A third group combines a relevance scalar with spreading over visual or cluster structure: MaxInfo~\cite{li2025maxinfo} selects a maximum-volume subset of visual embeddings; KTV~\cite{song2026ktv} clusters DINO features and picks representatives; SeViCES~\cite{sheng2026sevices} consolidates semantic-visual evidence; the ASCS baseline~\cite{li2026kfs} combines adaptive similarity with K-means cluster sampling; MDP3~\cite{Sun_2025_ICCV} and Q-Frame~\cite{Zhang_2025_ICCV} additionally wrap this third axis in a structured DPP-style kernel. Across all three groups the relevance representation and the spreading representation inhabit separate feature spaces. \textbf{FORGE} replaces this two-stage construction with a single query-conditioned geometry, on which relevance-weighted orthogonality of the frame embeddings carries both properties simultaneously. \\

Long-video understanding can be cast as an information-bottleneck problem. Regardless of how large the context window becomes, inference quality depends on the signal-to-noise ratio of query-relevant content in the input, which promotes the identification of that content to an optimization problem in its own right. The formal objective for this problem is the query-conditional mutual information between the selected subset and the answer, which is intractable at inference time in the absence of a joint distribution over frames, queries, and answers. A tractable surrogate is therefore required. The next section formalizes the objective and develops the surrogate we adopt.

\section{Method}
\label{sec:method}

\begin{figure*}[!htbp]
    \centering
    \includegraphics[width=0.95\textwidth]{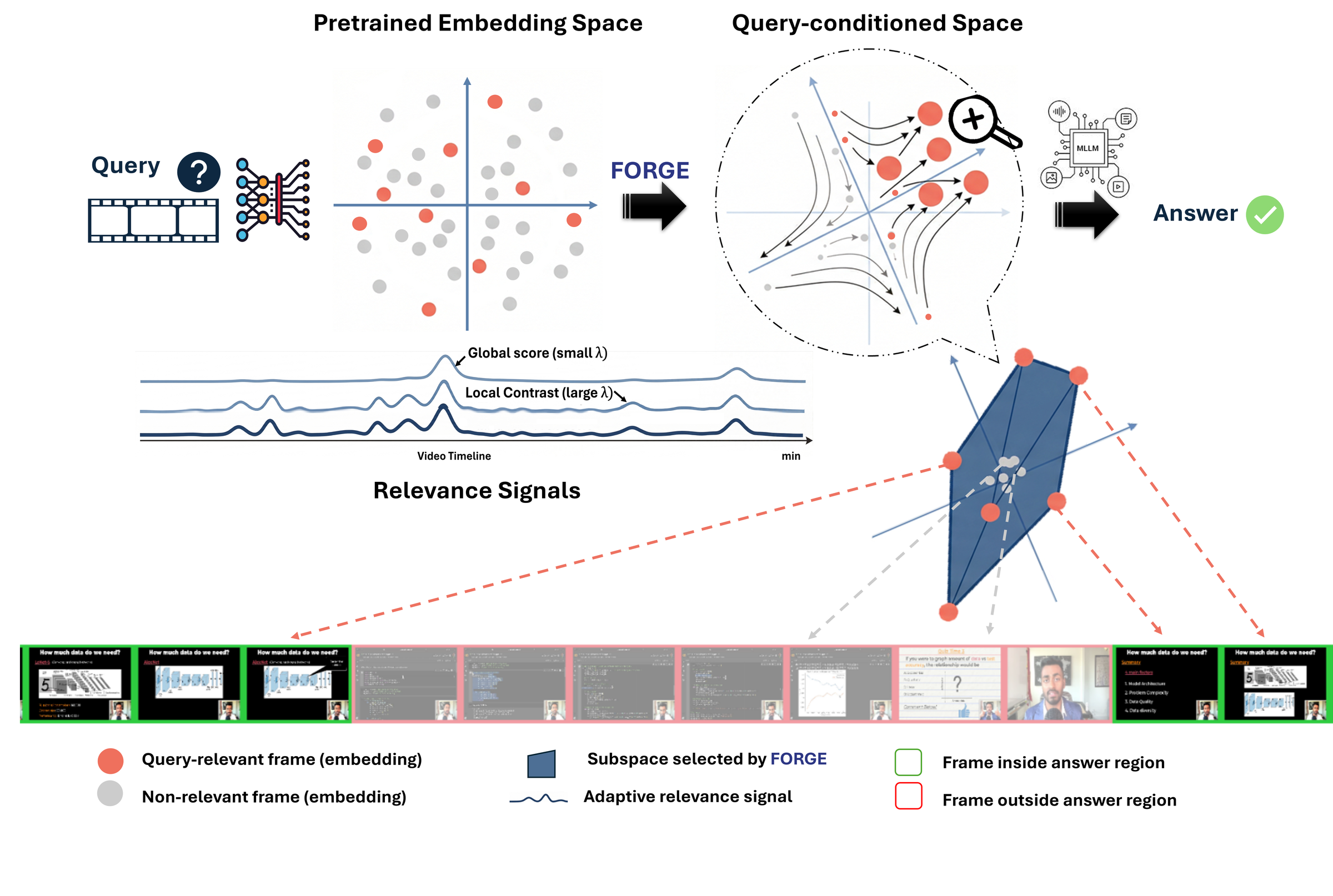}
   \caption{\textbf{The FORGE geometry.} FORGE induces a query-conditioned geometry on the pretrained frame embeddings, separating query-relevant frames along query-aligned directions while collapsing non-relevant frames toward the origin. Selecting the $K$ frames that span the maximum volume in this geometry unifies relevance and diversity into a single objective, and the selected frames contain the query-relevant content the MLLM needs to answer correctly. The bottom panel shows an example selection; green-bordered frames fall inside the annotated answer region.}
    \label{fig:Fig_method}
\end{figure*}


A video is represented as a sequence of $N$ frames indexed by $t \in \{1, \ldots, N\}$, paired with a text query $Q$. Each frame has a $d$-dimensional visual embedding $e_t \in \mathbb{R}^d$ produced by a pretrained visual encoder, and we collect the embeddings into a matrix $E \in \mathbb{R}^{N \times d}$. A relevance score $s_t \in \mathbb{R}$ is computed for each frame against the query through a multimodal matching model that reads both the query text and the frame. We use $s \in \mathbb{R}^N$ for the full score vector. The selection task is to pick a subset $S \subseteq \{1, \ldots, N\}$ of size $|S| = K$ that is passed to the downstream MLLM in place of the full video, denoted $V_S$, with answer $A$. The selection objective is the query-conditional mutual information (MI), denoted $I(\cdot)$, between the selected subset and the answer:
\begin{equation}
    S^\star \;=\; \arg\max_{\substack{S \subseteq \{1,\ldots,N\} \\ |S| = K}} \; I(V_S;\, A \mid Q).
    \label{eq:mi}
\end{equation}

\subsection*{From mutual information to query-conditioned volume}
\label{sec:method-objective}

Computing $I(V_S; A \mid Q)$ directly requires a model of the joint distribution over frames, queries, and answers that is unavailable at inference time. Applying the definition of conditional mutual information~\citep{cover2006elements} to Eq.~\eqref{eq:mi} gives
\begin{equation}
    I(V_S;\, A \mid Q) \;=\; H(A \mid Q) \;-\; H(A \mid V_S,\, Q).
    \label{eq:mi-decomp}
\end{equation}
The first term, $H(A \mid Q)$, is the inherent uncertainty of the answer given only the query and is constant with respect to $S$. Maximizing mutual information therefore reduces to minimizing $H(A \mid V_S, Q)$, that is, selecting the subset that makes the answer most predictable. Two complementary properties follow: each selected frame should be individually relevant to the query, and the selected frames together should span the query-relevant content rather than concentrate on one instance of it.
The remaining term is intractable without access to the downstream model, and \textbf{FORGE} reaches a computable objective in three steps. Selection is performed on frame embeddings rather than on raw frames; by the data processing inequality $I(f(V_S); A \mid Q) \leq I(V_S; A \mid Q)$ for any deterministic feature map $f$, so the embedding-space objective is a lower bound on Eq.~\eqref{eq:mi} and \textbf{FORGE} maximizes the bound. Answer uncertainty is taken to decrease as the selected frames carry more independent query-relevant content, which puts the entropy of the selected features in place of the entropy of the answer. Modeling the query-conditioned embeddings as Gaussian, that entropy is an affine function of the log-determinant of the subset's Gram matrix. The objective becomes the log-volume a subset spans in a geometry conditioned on the query.
 
The log-volume objective is submodular and, in its standard regularized form, monotone, so greedy selection carries the classical $(1-1/e)$ approximation guarantee~\citep{nemhauser1978analysis,kulesza2012determinantal}. It also inherits the diminishing-returns behavior of conditional mutual information: a frame redundant with those already selected yields near-zero marginal gain in either objective. Because the geometry is conditioned on the query, the diversity the objective rewards is diversity of \emph{query-relevant} content rather than visual diversity in general.
 
Concretely, a subset of frames that are all relevant but near-duplicates spans a small volume because the warped embeddings point in similar directions. A subset of diverse frames that are not query-relevant spans a small volume because the warping has collapsed them near the origin. The volume grows only when the two properties hold jointly.

\subsection*{FORGE}
\label{sec:method-forge}

The methods surveyed in Section~\ref{sec:related-selection} treat relevance and diversity as separate objectives optimized in disjoint feature spaces. A scalar score ranks individual frames, while a query-agnostic axis (temporal position, visual clusters, or embedding distance) spreads the selection. \textbf{FORGE} resolves scalar collapse by folding the relevance signal into the embedding geometry itself, so that a single volume objective in the resulting space simultaneously favors relevant frames and penalizes redundancy among them. The construction, illustrated in Figure~\ref{fig:Fig_method}, is a single forward computation: the raw relevance scores are refined into frame weights that induce a query-conditioned embedding space, and the $K$ frames spanning the largest volume in this geometry are read off by greedy orthogonalization.

\paragraph{Relevance as a scale-adaptive signal.}
The geometry is driven by a relevance signal that reads both the global score $s$ and its local structure. For a window of half-width $W$ around frame $t$, the local contrast $\mathrm{lc}_t = s_t - \mathrm{mean}(s_{t-W:t+W})$ measures the sharpness with which a frame stands out against its temporal neighborhood. The two signals capture different information: the global score isolates a distinctive visual cue when one exists, and the local contrast surfaces structure when the global landscape is flat, as happens for queries without a sharp cue or for videos where many frames look alike. \textbf{FORGE} combines them with
 
\begin{equation}
    \tilde{s}_t \;=\; s_t + \lambda \cdot \mathrm{lc}_t, \; \lambda \;=\; \mathrm{clip}\!\left(\frac{c}{\mathrm{std}(s) + \epsilon},\; \lambda_{\min},\; \lambda_{\max}\right),
    \label{eq:multires}
\end{equation}
where $\lambda$ is read off the score distribution of the current video and query rather than tuned. A high-variance global score yields a small $\lambda$ and the global signal carries the selection; a flat global score yields a large $\lambda$ and the local contrast is surfaced. The adaptive choice of $\lambda$ follows the principle of automatic scale selection, in which the signal itself indicates the informative scale rather than the scale being fixed in advance~\citep{lindeberg1998feature}, applied here to two scales rather than a continuum of resolutions. The combined score is then passed through a mean-centered sigmoid
 
\begin{equation}
    w_t \;=\; \mathrm{sigmoid}\!\left(\tau \cdot \frac{\tilde{s}_t - \mathrm{mean}(\tilde{s})}{\mathrm{std}(\tilde{s})}\right),
    \label{eq:sigmoid}
\end{equation}
 
which produces a frame weight $w_t \in (0,1)$ that preserves the absolute differences introduced by the scale-adaptive blend rather than compressing them to a fixed range. The constants $c$, $\lambda_{\min}$, $\lambda_{\max}$, $W$, $\tau$, and $\epsilon$ are fixed implementation constants shared across all datasets, queries, and MLLMs, and will be released with the public code.

\paragraph{A query-conditioned geometry.}
The frame weights induce a query-conditioned transformation of the visual embedding space
\begin{equation}
    \tilde{E} \;=\; \mathrm{diag}(w) \cdot E,
    \label{eq:warp}
\end{equation}
so that frames with higher query relevance point farther from the origin and frames with low relevance collapse toward it. The scaling leaves each embedding's direction unchanged; the angular structure is reshaped by the projection that follows. Because high-weight frames dominate the singular value decomposition of $\tilde{E}$, the top singular vectors align with the directions that query-relevant frames span. Projecting all frames onto that subspace discards the directions in which irrelevant frames were distinctive, collapsing their angular separation, while relevant frames that differ along query-relevant dimensions retain their spread. Volume maximization in the projected geometry therefore rewards diversity of query-relevant content rather than visual diversity in general.
 
To work in the intrinsic geometry of this warped space rather than its ambient coordinates, \textbf{FORGE} takes the singular values $\sigma_i$ of $\tilde{E}$ and reads them as a distribution $\hat{\sigma}_i = \sigma_i / \sum_j \sigma_j$. The spectral entropy $H_{\sigma} = -\sum_i \hat{\sigma}_i \log \hat{\sigma}_i$ has perplexity $\exp(H_{\sigma})$, a standard estimate of effective rank, and the projection dimension is set to $R = \mathrm{round}(\exp(H_{\sigma}))$. Projecting $\tilde{E}$ onto the subspace spanned by its top-$R$ right singular vectors yields
\begin{equation}
    \tilde{E}_{\mathrm{proj}} \in \mathbb{R}^{N \times R},
    \label{eq:proj}
\end{equation}
the feature space in which the selection is performed. The dimension $R$ is determined by the data: videos whose warped embeddings concentrate along a few directions (a short event, a single speaker) are projected low-dimensional, while videos whose warped embeddings spread across many directions retain more dimensions.
 
\paragraph{Selection as volume maximization.}
In the projected geometry, we identify the subset that is jointly relevant and non-redundant with the subset whose warped embeddings span the largest volume. \textbf{FORGE} reads this subset greedily. Starting from $Z^{(0)} = \tilde{E}_{\mathrm{proj}}$ and $S = \emptyset$, each iteration selects
\begin{equation}
    i^\star \;=\; \arg\max_i \, \| Z^{(k)}_i \|^2,
    \label{eq:greedy}
\end{equation}
appends $i^\star$ to $S$, and updates the residuals by subtracting the component of every row along $Z^{(k)}_{i^\star}$. Each iteration therefore picks the frame whose residual is most orthogonal to the span of the frames already chosen. Eq.~\eqref{eq:greedy} is greedy ascent on the log-volume objective, since appending a frame raises the log-determinant of the Gram matrix by the log of its squared residual norm; the selection therefore carries the $(1-1/e)$ guarantee stated above. Algorithm~\ref{alg:forge} summarizes the full procedure: a single pass through the scores and embeddings, no training, and no per-video tuning.
 
\begin{algorithm}[t]
\caption{FORGE}
\label{alg:forge}
\begin{algorithmic}[]
\Require Embeddings $E \in \mathbb{R}^{N \times d}$, relevance scores $s \in \mathbb{R}^N$, budget $K$
\Ensure Selected frame set $S$, $|S| = K$
\State Compute scale-adaptive score $\tilde{s}$ and frame weights $w$
\State Form the query-conditioned geometry $\tilde{E}_{\mathrm{proj}}$
\State $S \gets \emptyset$, $Z \gets \tilde{E}_{\mathrm{proj}}$
\For{$k = 1$ to $K$}
    \State $i^\star \gets \arg\max_i \| Z_i \|^2$, $\; S \gets S \cup \{i^\star\}$
    \State Orthogonalize $Z$ against $Z_{i^\star}$
\EndFor
\State \Return $S$
\end{algorithmic}
\end{algorithm}

\section{Experiments}
\label{sec:experiments}

We present our experimental setup, followed by the main results on selection quality and downstream video question answering, an evaluation across MLLMs, an ablation of \textbf{FORGE}'s mechanisms, an analysis of the properties of the selection approaches we evaluate, and an evaluation of cost efficiency.

\subsection*{Setup}
\label{sec:exp-setup}

\textbf{Datasets and metrics.} We evaluate on Video-MME~\citep{fu2025video} and LongVideoBench~\citep{wu2024longvideobench}, two long-form videoQA benchmarks. Video-MME includes 900 videos totaling 254 hours with 2,700 human-annotated questions, while LongVideoBench consists of 3,763 videos spanning over 760 hours with 6,678 multiple-choice questions. We report selection quality with Keyframe Recall (KFR), Scene Hit Rate (SHR), and the Unified Keyframe Selection Score (UKSS) introduced by~\citet{li2026kfs}, and downstream performance with top-1 accuracy. All experiments use frame budgets $K \in \{16, 32, 64\}$.

\textbf{Models and baselines.} We evaluate \textbf{FORGE} against an exhaustive set of recent state-of-the-art training-free methods covering the approaches surveyed in Sec.~\ref{sec:related-selection}~\citep{tang2025adaptive,li2026kfs,Sun_2025_ICCV,li2025maxinfo,son2024csta}, with uniform sampling as the floor baseline. Downstream video understating is assessed on a comprehensive collection of open-source MLLMs across multiple model families and parameter scales, with the frame pool, decoding settings, and input preprocessing fixed across all experiments so that only the selection method varies. The specific baselines and MLLMs used in each experiment are listed in the corresponding table. Training-based and language-model-in-the-loop methods are not directly comparable in cost (Sec.~\ref{sec:related-selection},~\ref{sec:related-agentic}) and are not included in this comparison.

\textbf{Implementation.} All methods, including \textbf{FORGE}, sample from a 1-FPS pool from each video, and use the same pretrained visual encoder for frame embeddings and the same pretrained multimodal encoder for query relevance scores. Each method's parameters are fixed across both datasets and all budgets, with no per-video tuning. All baselines are reproduced from their public code or published papers.
\textbf{Reproducibility.} \textbf{FORGE} is deterministic given a video, query, and frame budget, and all MLLMs are evaluated with fixed decoding settings, so repeated runs yield identical selections and identical answers. All reported numbers were confirmed over three independent runs.

\begin{table}[!htbp]
\centering
\renewcommand{\arraystretch}{1.3}
\resizebox{\columnwidth}{!}{%
\begin{tabular}{l *{9}{c}}
\toprule
\multirow{2}{*}{Method} & \multicolumn{9}{c}{\textbf{Video-MME}} \\ \cmidrule(l){2-10}
& \multicolumn{3}{c}{$K{=}16$} & \multicolumn{3}{c}{$K{=}32$} & \multicolumn{3}{c}{$K{=}64$} \\
\cmidrule(lr){2-4} \cmidrule(lr){5-7} \cmidrule(lr){8-10}
& KFR & SHR & UKSS & KFR & SHR & UKSS & KFR & SHR & UKSS \\
\midrule
Uniform  & .182 & .618 & .380 & .178 & .714 & .401 & .163 & .811 & .407 \\
AKS~\citep{tang2025adaptive}      & .216 & \underline{.749} & \underline{.441} & .208 & .766 & .442 & .201 & .830 & .451 \\
ASCS~\citep{li2026kfs}     & \underline{.250} & .649 & .439 & \underline{.229} & .755 & \underline{.455} & .198 & .801 & .436 \\
MDP3~\citep{Sun_2025_ICCV}     & .220 & .677 & .432 & .209 & \underline{.773} & .445 & \underline{.204} & \underline{.846} & \underline{.455} \\
MaxInfo~\citep{li2025maxinfo}  & .167 & .588 & .358 & .166 & .705 & .387 & .155 & .772 & .388 \\
\rowcolor{gray!15}
\textbf{FORGE} & \textbf{.356} & \textbf{.758} & \textbf{.551} & \textbf{.385} & \textbf{.834} & \textbf{.583} & \textbf{.415} & \textbf{.886} & \textbf{.608} \\
\bottomrule \\
\multirow{2}{*}{} & \multicolumn{9}{c}{\textbf{LongVideoBench}} \\ \cmidrule(l){2-10}
& \multicolumn{3}{c}{$K{=}16$} & \multicolumn{3}{c}{$K{=}32$} & \multicolumn{3}{c}{$K{=}64$} \\
\cmidrule(lr){2-4} \cmidrule(lr){5-7} \cmidrule(lr){8-10}
& KFR & SHR & UKSS & KFR & SHR & UKSS & KFR & SHR & UKSS \\
\midrule
Uniform  & .073 & .326 & .184 & .067 & .443 & .207 & .059 & .565 & .222 \\
AKS~\citep{tang2025adaptive}      & .130 & \underline{.586} & .312 & .101 & .590 & .286 & .082 & .671 & .284 \\
ASCS~\citep{li2026kfs}     & \underline{.168} & .564 & \underline{.347} & .140 & .650 & .344 & .111 & .729 & .331 \\
MDP3~\citep{Sun_2025_ICCV}     & .157 & .571 & .336 & \underline{.152} & \underline{.674} & \underline{.354} & \underline{.151} & \underline{.783} & \underline{.372} \\
MaxInfo~\citep{li2025maxinfo}  & .069 & .336 & .185 & .065 & .436 & .206 & .057 & .555 & .219 \\
\rowcolor{gray!15}
\textbf{FORGE} & \textbf{.272} & \textbf{.681} & \textbf{.458} & \textbf{.272} & \textbf{.762} & \textbf{.472} & \textbf{.271} & \textbf{.832} & \textbf{.482} \\
\bottomrule
\end{tabular}%
}
\caption{Selection quality on Video-MME (top) and LongVideoBench (bottom) at $K \in \{16, 32, 64\}$, reported as Keyframe Recall (KFR), Scene Hit Rate (SHR), and the Unified Keyframe Selection Score (UKSS) of~\citet{li2026kfs}. FORGE rows are highlighted; best results are \textbf{bolded}, second-best \underline{underlined}. FORGE leads every competitor on UKSS at every budget on both benchmarks.}
\label{tab:selection}
\end{table}

\subsection{Main results: selection quality and downstream accuracy}
\label{sec:exp-main}

\textbf{FORGE} improves both selection quality and downstream accuracy on video question answering with MLLMs, across both benchmarks and all three frame budgets. Selection quality is computed using ground-truth keyframe and scene annotations in Table~\ref{tab:selection}, and downstream top-1 VQA accuracy for two primary MLLMs is reported in Table~\ref{tab:vqa-primary}.

Table~\ref{tab:selection} shows that \textbf{FORGE}'s KFR gain over baselines is larger than its SHR gain. SHR rewards selecting any frame in a related scene, so any method that distributes its picks across the video timeline has a higher probability of hitting a scene-relevant region. KFR, in contrast, rewards selecting more of the specific frames annotated as keyframes within those scenes, and \textbf{FORGE} consistently picks more keyframes while keeping a high SHR. This indicates that the additional frames \textbf{FORGE} selects at larger budgets continue to be scene-relevant and inside the annotated keyframe regions rather than spread to irrelevant regions on the timeline, despite the sparsity of answer-relevant content in long videos.

\begin{table}[!htbp]
\centering
\renewcommand{\arraystretch}{1.2}
\resizebox{0.9\columnwidth}{!}{%
\begin{tabular}{@{}p{1.3cm}lcccc@{}}
\toprule
\multicolumn{2}{c}{\multirow{2}{*}{MLLM}} & \multirow{2}{*}{Method} & \multicolumn{3}{c}{\textbf{Video-MME}} \\ \cmidrule(l){4-6}
\multicolumn{2}{c}{} && $K{=}16$ & $K{=}32$ & $K{=}64$ \\ \midrule
\multirow{7}{1.3cm}{Qwen2.5-VL 7B~\citep{bai2025qwen2}}
 &  & Uniform  & 56.00\% & 58.22\% & 60.67\% \\
 &  & AKS~\citep{tang2025adaptive}      & \underline{59.68\%} & 59.80\% & 61.16\% \\
 &  & ASCS~\citep{li2026kfs}     & 57.31\% & 58.59\% & 59.61\% \\
 &  & MDP3~\citep{Sun_2025_ICCV}     & 57.02\% & \underline{60.32\%} & \underline{61.26\%} \\
 &  & MaxInfo~\citep{li2025maxinfo}  & 56.79\% & 58.17\% & 59.82\% \\
 &  & CSTA~\citep{son2024csta}     & 58.00\% & 59.20\% & -- \\
\rowcolor{gray!15}
 &  & \textbf{FORGE} & \textbf{60.56\%} & \textbf{62.82\%} & \textbf{63.49\%} \\ \midrule
\multirow{7}{1.3cm}{Gemma3 12B~\citep{gemma32025}}
 &  & Uniform  & 59.05\% & 62.35\% & 63.26\% \\
 &  & AKS~\citep{tang2025adaptive}      & \underline{63.58\%} & \underline{63.44\%} & 63.96\% \\
 &  & ASCS~\citep{li2026kfs}     & 60.94\% & 63.09\% & 63.25\% \\
 &  & MDP3~\citep{Sun_2025_ICCV}     & 61.79\% & 62.86\% & \underline{65.42\%} \\
 &  & MaxInfo~\citep{li2025maxinfo}  & 61.96\% & 63.38\% & 63.52\% \\
 &  & CSTA~\citep{son2024csta}     & -- & -- & -- \\
\rowcolor{gray!15}
 &  & \textbf{FORGE} & \textbf{65.52\%} & \textbf{67.34\%} & \textbf{67.91\%} \\ \bottomrule \\
\multicolumn{2}{c}{\multirow{2}{*}{MLLM}} & \multirow{2}{*}{Method} & \multicolumn{3}{c}{\textbf{LongVideoBench}} \\ \cmidrule(l){4-6}
\multicolumn{2}{c}{} && $K{=}16$ & $K{=}32$ & $K{=}64$ \\ \midrule
\multirow{7}{1.3cm}{Qwen2.5-VL 7B~\citep{bai2025qwen2}}
 &  & Uniform  & 53.89\% & 56.89\% & 58.81\% \\
 &  & AKS~\citep{tang2025adaptive}      & 55.64\% & 54.15\% & 56.44\% \\
 &  & ASCS~\citep{li2026kfs}     & \underline{57.07\%} & \underline{58.84\%} & \underline{60.13\%} \\
 &  & MDP3~\citep{Sun_2025_ICCV}     & 56.15\% & 55.89\% & 57.59\% \\
 &  & MaxInfo~\citep{li2025maxinfo}  & 47.87\% & 49.15\% & 50.37\% \\
 &  & CSTA~\citep{son2024csta}     & 38.2\% & 39.3\% & 38.7\% \\
\rowcolor{gray!15}
 &  & \textbf{FORGE} & \textbf{60.25\%} & \textbf{63.14\%} & \textbf{63.66\%} \\ \midrule
\multirow{7}{1.3cm}{Gemma3 12B~\citep{gemma32025}}
 &  & Uniform  & 51.35\% & 52.18\% & \underline{56.17\%} \\
 &  & AKS~\citep{tang2025adaptive}      & 52.86\% & 52.81\% & 53.35\% \\
 &  & ASCS~\citep{li2026kfs}     & \underline{53.94\%} & 53.35\% & 54.58\% \\
 &  & MDP3~\citep{Sun_2025_ICCV}     & 52.80\% & \underline{54.75\%} & 54.93\% \\
 &  & MaxInfo~\citep{li2025maxinfo}  & 46.93\% & 50.25\% & 49.16\% \\
 &  & CSTA~\citep{son2024csta}     & 39.4\% & 40.2\% & 37.8\% \\
\rowcolor{gray!15}
 &  & \textbf{FORGE} & \textbf{58.18\%} & \textbf{59.94\%} & \textbf{61.36\%} \\ \bottomrule
\end{tabular}%
}
\caption{ Top-1 VQA accuracy on Video-MME (top) and LongVideoBench (bottom)
at $K \in \{16, 32, 64\}$ for Qwen2.5-VL 7B and Gemma3 12B.
FORGE rows are highlighted; best results are \textbf{bolded}, second-best \underline{underlined}.
FORGE is the top method on every MLLM, benchmark, and budget combination.}
\label{tab:vqa-primary}
\end{table}

As shown in Table~\ref{tab:vqa-primary}, \textbf{FORGE} outperforms all baselines by a larger margin on LongVideoBench, where videos are longer on average and answer regions are narrower. Despite \textbf{FORGE}'s consistent improvement over uniform sampling at every budget, several baselines perform worse than it with more frames. This pattern reveals a distinction between similarity and relevance. Methods that rank frames by pointwise similarity to the query select frames that \emph{look like} the query, but not necessarily content that is relevant to the answer.

\textit{\noindent\textbf{Takeaway.} When the similarity-selected frames do not overlap with the relevant content, the result is worse than query-agnostic uniform sampling. Preserving the query's high-dimensional structure in the selection geometry, rather than collapsing it to a scalar similarity score, yields frames that contain the information the MLLM needs to answer correctly.}

\begin{figure*}[!htbp]
    \centering
    \begin{subfigure}[t]{\linewidth}
        \centering
        \includegraphics[width=\linewidth]{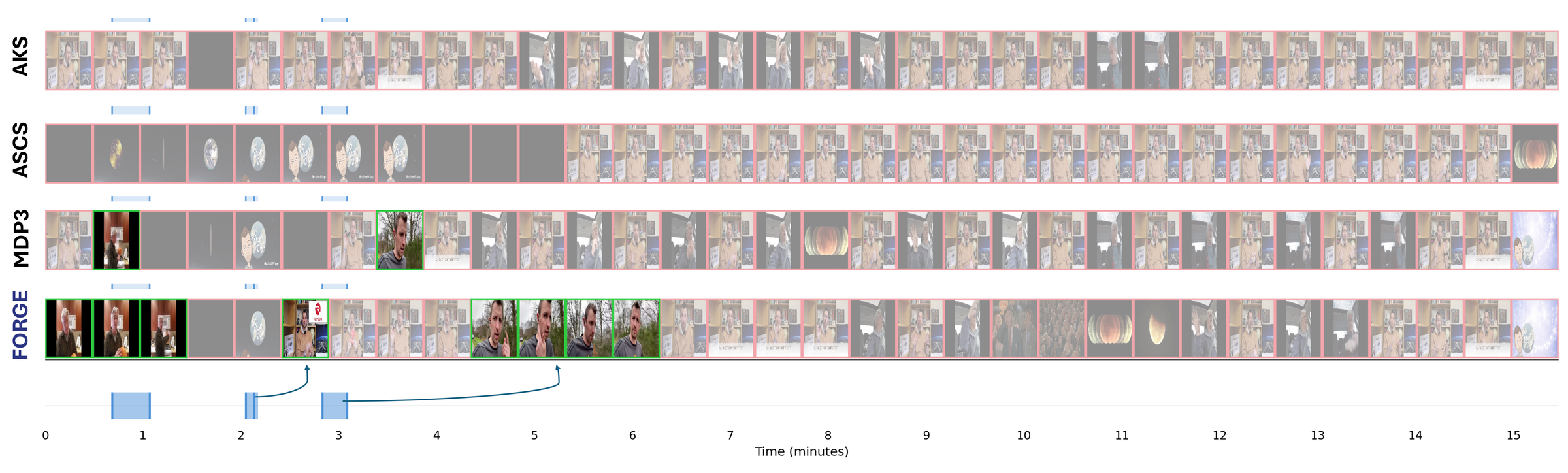}
    \caption{LongVideoBench, 15-min video.}
        \label{fig:qual-lvb}
    \end{subfigure}


    \begin{subfigure}[t]{\linewidth}
        \centering
        \includegraphics[width=\linewidth]{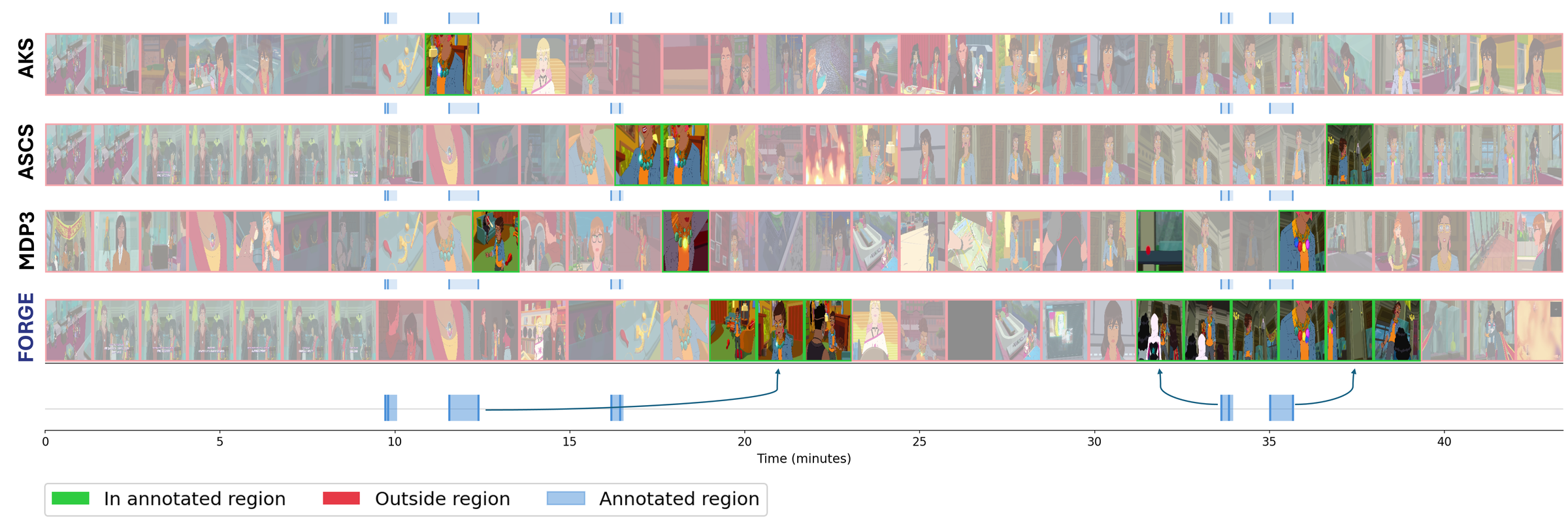}
        \caption{Video-MME, 43-min video.}
        \label{fig:qual-videomme}
    \end{subfigure}

    \caption{Qualitative comparison of KFR and SHR on a Video-MME and a LongVideoBench Video at $K{=}32$, illustrating the KFR margin reported in Table~\ref{tab:selection}. Each row shows the 32 thumbnails selected by one method, with green borders for frames inside the annotated answer regions and red borders for frames outside. Blue markers along the timeline shows the ground-truth answer regions. FORGE selects more frames inside the annotated regions, where the competing selectors find it challenging to find key frames.}
    \label{fig:qualitative}
\end{figure*}

\begin{figure*}[!htbp]
    \centering
    \begin{subfigure}[t]{0.9\linewidth}
        \centering
        \includegraphics[width=\linewidth]{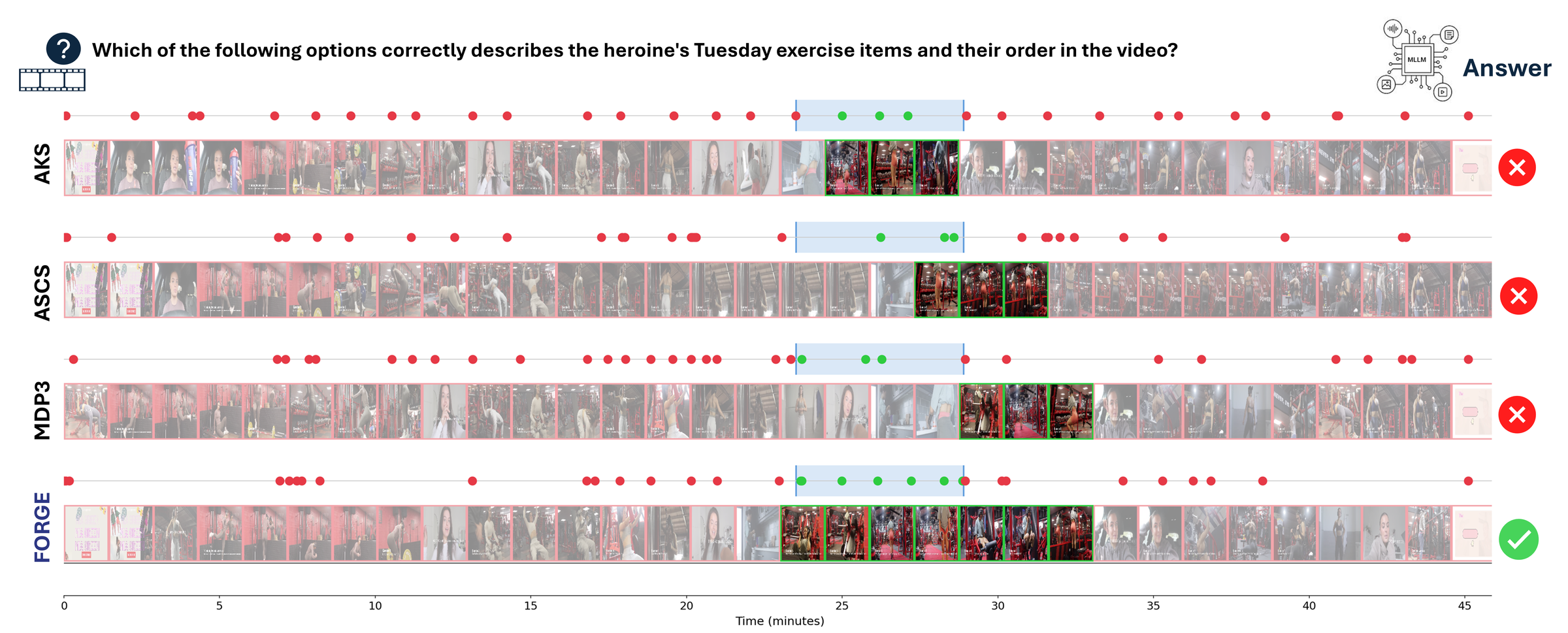}
        \caption{Video-MME, 46-min video.}
        \label{fig:qual-vqa-a}
    \end{subfigure}


    \begin{subfigure}[t]{0.9\linewidth}
        \centering
        \includegraphics[width=\linewidth]{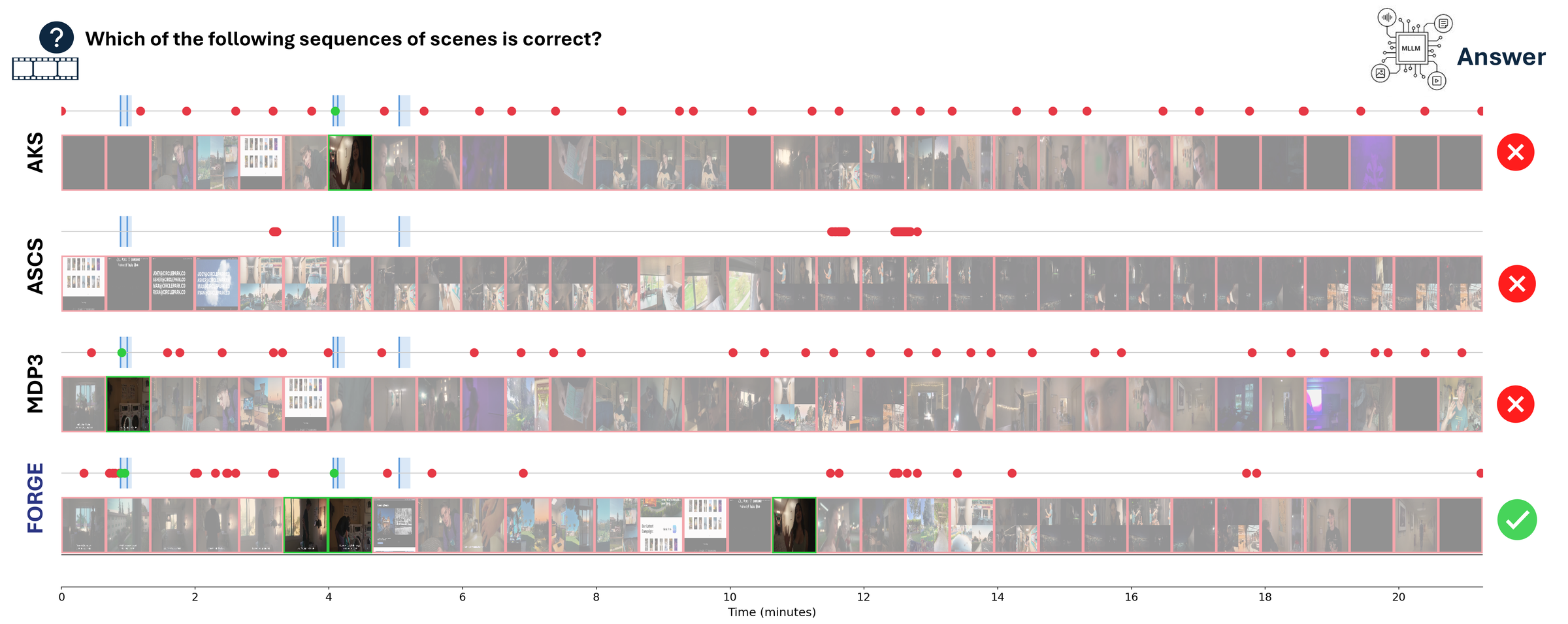}
        \caption{LongVideoBench, 16-min video.}
        \label{fig:qual-vqa-b}
    \end{subfigure}
    \begin{subfigure}[t]{0.9\linewidth}
        \centering
        \includegraphics[width=\linewidth]{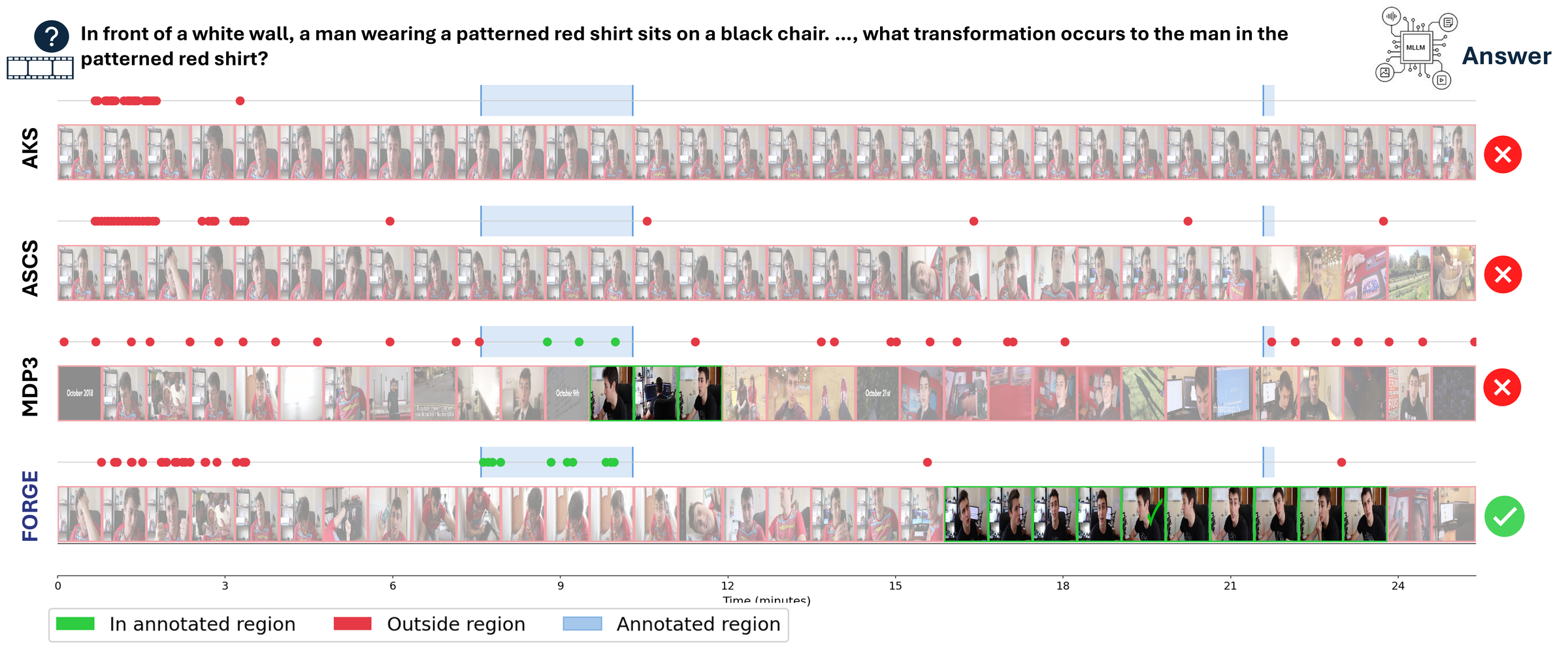}
        \caption{LongVideoBench, 25-min video.}
        \label{fig:qual-vqa-c}
    \end{subfigure}

    \caption{Qualitative comparison of MLLM answers between FORGE and leading competing selectors at $K{=}32$. Blue shaded bands represent annotated answer regions; dots above each row represent selected frames. Competing methods select more irrelevant frame and the MLLM answers incorrectly.}
    \label{fig:qualitative-vqa}
\end{figure*}

\subsection{Generalization across MLLM backbones}
\label{sec:exp-generalization}

To assess whether the improvements are specific to a particular MLLM model, we evaluate \textbf{FORGE} on additional open-source MLLMs spanning different architectural categories and a 4B--32B parameter range; results are reported in Table~\ref{tab:vqa-extra}.
The improvement over uniform sampling is stable across standard instruction-tuned models at all parameter scales, and extends to long-video specialized backbones such as LongVA and SeViLA, which process frames differently from standard MLLMs. The accuracy advantage is preserved even for token-compression backbones, where the MLLM itself suppresses redundant visual tokens, suggesting that frame-level selection and token-level compression are complementary rather than redundant.

\begin{table}[tbp]
\centering
\resizebox{\columnwidth}{!}{%
\begin{tabular}{@{}llcccccc@{}}
\toprule
\multirow{2}{*}{MLLM} & \multirow{2}{*}{Method} & \multicolumn{3}{c}{\textbf{Video-MME}} & \multicolumn{3}{c}{\textbf{LongVideoBench}} \\ \cmidrule(l){3-8}
 &  & $K{=}16$ & $K{=}32$ & $K{=}64$ & $K{=}16$ & $K{=}32$ & $K{=}64$ \\ \midrule
\multirow{2}{*}{LLaVA-OneVision 7B~\citep{li2024llava}} & Uniform & 56.04\% & 57.44\% & 58.26\% & 56.59\% & 56.29\% & 58.96\% \\
 & \textbf{FORGE} & \textbf{60.27\%} & \textbf{62.36\%} & \textbf{62.17\%} & \textbf{61.38\%} & \textbf{62.94\%} & \textbf{61.88\%} \\ \midrule
\multirow{2}{*}{Qwen2.5-VL 7B~\citep{bai2025qwen2}} & Uniform & 56.00\% & 58.22\% & 60.67\% & 53.89\% & 56.89\% & 58.81\% \\
 & \textbf{FORGE} & \textbf{60.56\%} & \textbf{62.82\%} & \textbf{63.49\%} & \textbf{60.25\%} & \textbf{63.14\%} & \textbf{63.66\%} \\ \midrule
\multirow{2}{*}{Gemma3 12B~\citep{gemma32025}} & Uniform & 59.05\% & 62.35\% & 63.26\% & 51.35\% & 52.18\% & 56.17\% \\
 & \textbf{FORGE} & \textbf{65.52\%} & \textbf{67.34\%} & \textbf{67.91\%} & \textbf{58.18\%} & \textbf{59.94\%} & \textbf{61.36\%} \\ \midrule
\multirow{2}{*}{Qwen2.5-VL 32B~\citep{bai2025qwen2}} & Uniform & 58.60\% & 60.48\% & 63.62\% & 55.99\% & 57.49\% & 59.49\% \\
 & \textbf{FORGE} & \textbf{62.42\%} & \textbf{63.47\%} & \textbf{66.87\%} & \textbf{61.59\%} & \textbf{64.67\%} & \textbf{64.41\%} \\ \midrule
\multicolumn{8}{@{}l}{\textit{Long-video specialized backbones}} \\ \midrule
\multirow{2}{*}{LongVA 7B~\citep{zhang2024long}} & Original (128 f) & \multicolumn{3}{c}{52.6\%} & \multicolumn{3}{c}{51.3\%} \\
& \textbf{FORGE (32 f)} & \multicolumn{3}{c}{\textbf{63.47\%}} & \multicolumn{3}{c}{\textbf{64.67\%}} \\ \midrule
\multirow{2}{*}{SeViLA 4B~\citep{yu2023self}} & Original &  & 39.5\% &  &  & 39.3\% &  \\
 & \textbf{FORGE} &  & \textbf{59.94\%} &  &  & \textbf{60.02\%} &  \\ \midrule
\multicolumn{8}{@{}l}{\textit{Token-compression backbones}} \\ \midrule
\multirow{2}{*}{InternVideo2.5-Chat 8B~\citep{wang2025internvideo25}} & Uniform & 57.26\% & 58.15\% & 61.78\% & 54.46\% & 56.36\% & 60.76\% \\
 & \textbf{FORGE} & \textbf{61.71\%} & \textbf{64.02\%} & \textbf{64.58\%} & \textbf{61.85\%} & \textbf{65.02\%} & \textbf{65.03\%} \\ \midrule
\multirow{2}{*}{InternVL2.5 26B~\citep{chen2024internvl25}} & Uniform & 62.89\% & 64.96\% & 66.04\% & 60.40\% & 62.20\% & 64.14\% \\
 & \textbf{FORGE} & \textbf{66.91\%} & \textbf{69.10\%} & \textbf{68.54\%} & \textbf{65.33\%} & \textbf{67.83\%} & \textbf{65.05\%} \\ \bottomrule
\end{tabular}%
}
\caption{Top-1 accuracy on additional MLLMs spanning a 4B--32B parameter range. Italic group labels indicate models with a specialized design for long-video or token compression. \textbf{FORGE} rows are in bold.}
\label{tab:vqa-extra}
\end{table}

\textit{\noindent\textbf{Takeaway.} Because \textbf{FORGE}'s selection is computed before any MLLM reads the frames, the consistent gain across all evaluated backbones is a property of the selected frames and not of any particular MLLM's capabilities, confirming that the method is model-agnostic.}

\subsection{Ablation: contribution of each component}
\label{sec:exp-ablation}

\begin{figure*}[htb]
    \centering
    \begin{subfigure}[t]{0.49\textwidth}
        \centering
        \includegraphics[width=\linewidth]{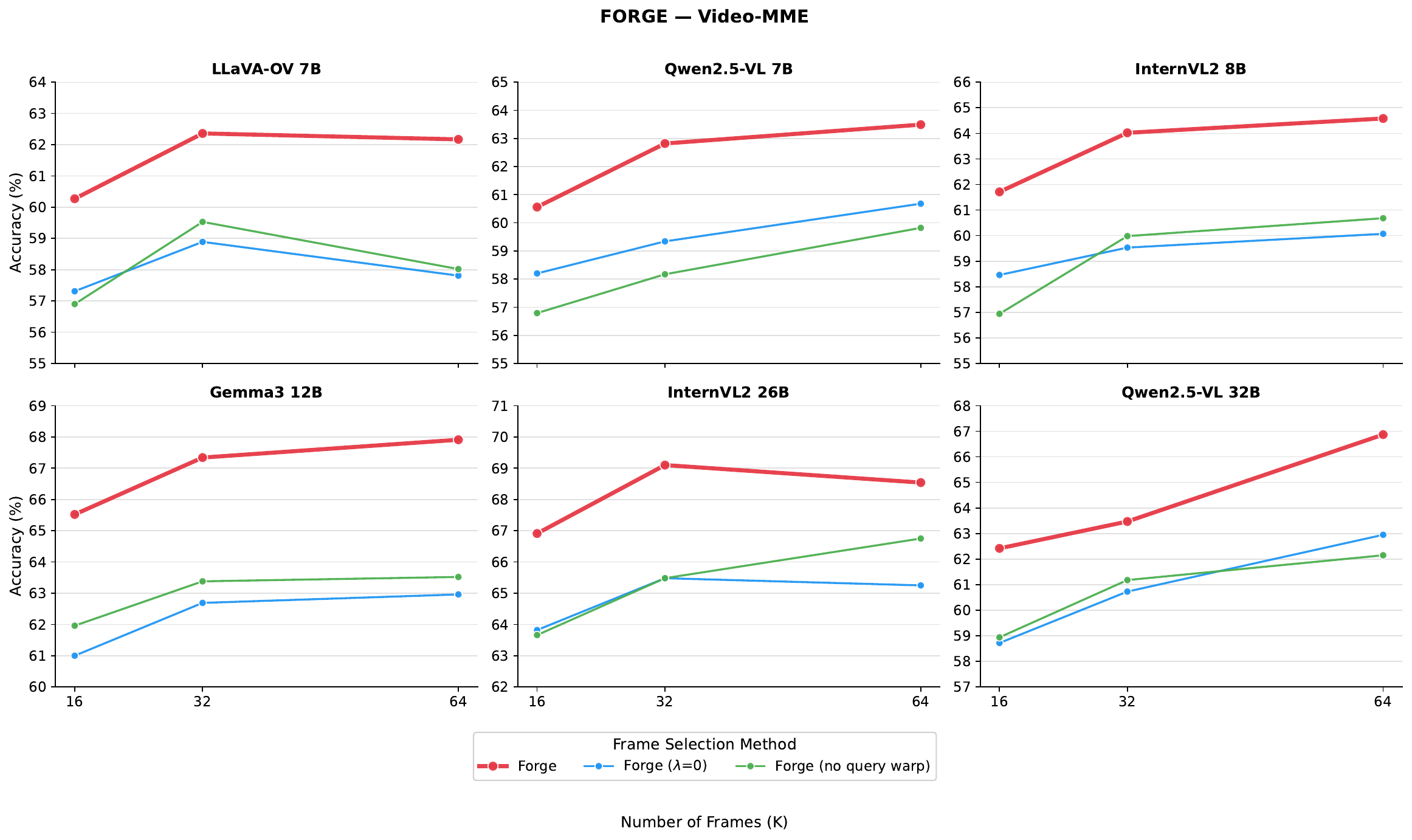}
        \label{fig:ablation-vmme}
    \end{subfigure}
    \begin{subfigure}[t]{0.49\textwidth}
        \centering
        \includegraphics[width=\linewidth]{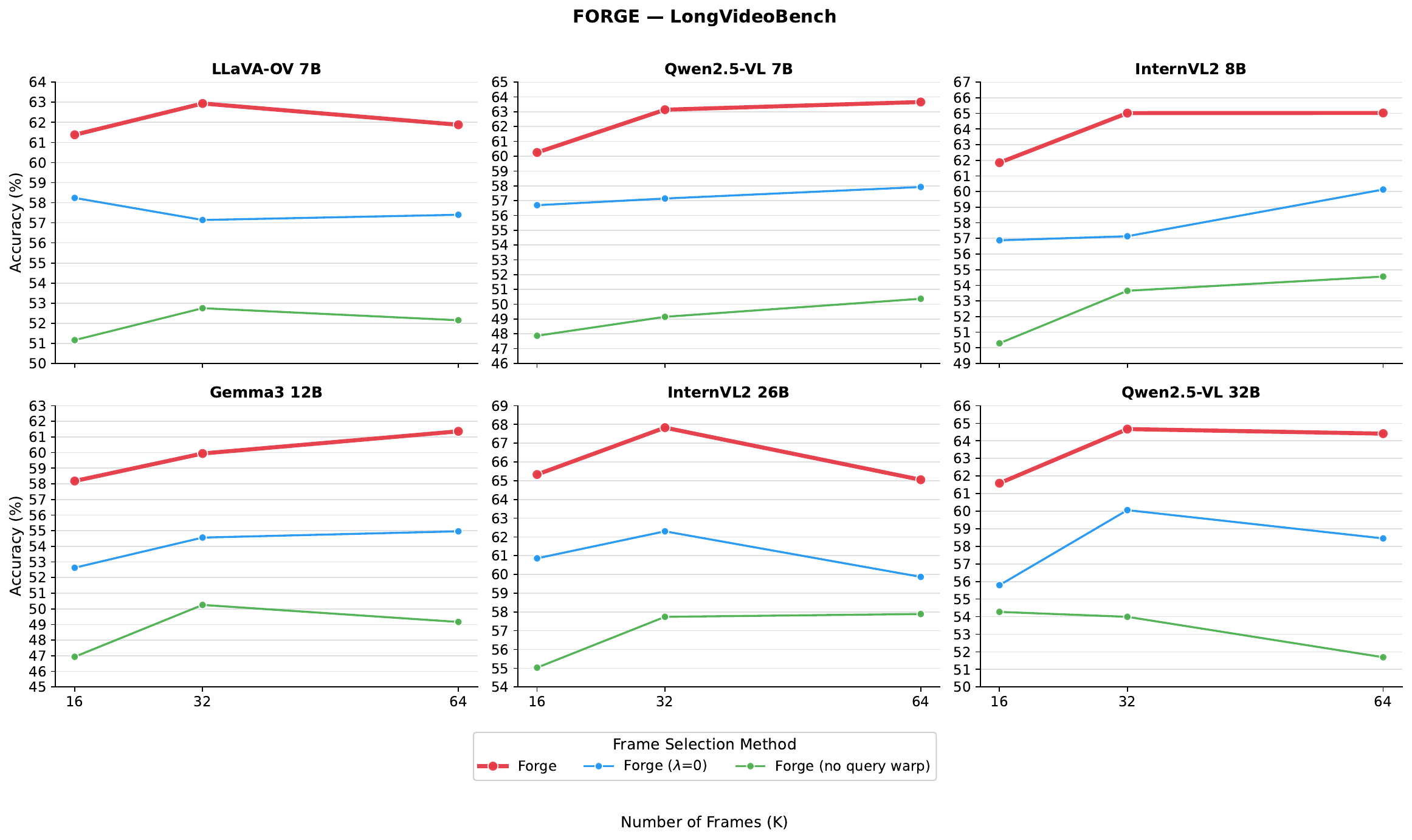}
        \label{fig:ablation-lvb}
    \end{subfigure}
    \caption{Component ablation of FORGE on six MLLMs across two benchmarks. Each panel plots top-1 accuracy as a function of frame budget $K \in \{16, 32, 64\}$ for full FORGE, FORGE ($\lambda{=}0$), and FORGE ($-$query warp). Full FORGE leads every partial variant in every panel on both benchmarks.}
    \label{fig:ablation-per-model}
\end{figure*}

\textbf{FORGE}'s selection is defined by a unified query-conditioned geometry in which frames are weighted by $w_t$, derived from both the global relevance score $s$ and the local contrast $\mathrm{lc}$ (Eq.~\ref{eq:multires}), and the selected subset is the one that spans the largest volume in this geometry. We ablate each component of $w_t$ individually: \textbf{FORGE ($\lambda{=}0$)} removes the local-contrast term, and \textbf{FORGE ( No query warp)} sets $w_t{=}1$, reducing selection to volume maximization on the pretrained embeddings. Figure~\ref{fig:ablation-per-model} reports the accuracy of each variant on six MLLMs across both benchmarks. Full \textbf{FORGE} outperforms both partial variants in every panel, with the query-conditioned geometry accounting for most of the accuracy improvement on LongVideoBench.

Figure~\ref{fig:ablation-per-model} shows that removing the query warp leads to a larger accuracy drop than zeroing the local-contrast term on both benchmarks, and on LongVideoBench pushes most MLLMs below uniform sampling. Without the warp, volume maximization spans the pretrained embedding space along directions the query does not select, and a selection composed of such frames is worse than uniform sampling when answer-relevant content is sparse. Removing the local-contrast term reduces accuracy but never below uniform sampling, as it refines an already query-conditioned signal rather than replacing it.

\textit{\noindent\textbf{Takeaway.} Removing the query-conditioned weighting leads to a larger accuracy drop than removing the local-contrast term, showing that the query-conditioned geometry is what lets a single selection objective optimize relevance and diversity jointly.}

\subsection{Properties of the selected frames}
\label{sec:exp-character}

We analyze three properties of the frame sets selected by each method: visual redundancy, temporal distribution relative to uniform sampling, and consistency of the selection under a budget increase, reported in Table~\ref{tab:properties}. These properties are in tension with one another, and each competing method performs well on one or two at the expense of the others.

\begin{table}[t]
\centering
\resizebox{\columnwidth}{!}{%
\begin{tabular}{@{}lccc@{}}
\toprule
Method & Pairwise Cosine $\downarrow$ & Wasserstein & Jaccard $16{\to}32$ \\
\midrule
AKS~\citep{tang2025adaptive}      & 0.7260 & 0.0236 & 0.6610 \\
ASCS~\citep{li2026kfs}            & 0.6952 & 0.0848 & 0.8133 \\
MDP3~\citep{Sun_2025_ICCV}        & 0.6848 & 0.0562 & 0.5518 \\
MaxInfo~\citep{li2025maxinfo}     & 0.5772 & 0.0671 & 0.5000 \\
\textbf{FORGE}                    & \textbf{0.6937} & \textbf{0.1096} & \textbf{0.5000} \\
\midrule
Top-k (reference)                 & 0.7582 & 0.1425 & 0.5000 \\
Uniform (reference)               & 0.6809 & --     & 0.8807 \\
\bottomrule
\end{tabular}%
}
\caption{Properties of the selected frame sets at $K=32$ on Video-MME. Pairwise cosine similarity measures visual redundancy (lower is less redundant). Wasserstein distance to the uniform distribution on $[0,1]$ measures how query-conditional the temporal placement is (values near zero indicate near-uniform, query-independent placement). Jaccard distance between the $K=16$ and $K=32$ selections measures selection consistency under a budget increase (0.5 corresponds to exact containment of the smaller set in the larger one). Top-k and Uniform are included as reference points for query-driven temporal concentration and for query-independent sampling.}
\label{tab:properties}
\end{table}

The metrics reported in Table~\ref{tab:properties} infdicate that AKS, MDP3, and MaxInfo produce near-uniform temporal distributions, with MaxInfo additionally achieving the lowest visual redundancy. ASCS's $K{=}16$ and $K{=}32$ selections are largely disjoint, and Top-k combines the highest visual redundancy with the most query-concentrated temporal distribution. \textbf{FORGE} achieves a balanced profile across all three properties: it is temporally query-conditioned, its $K{=}16$ selection is exactly contained in its $K{=}32$ selection, and its visual redundancy sits in the middle of the range, below the query-concentrated Top-k and AKS selections and above the redundancy-minimizing MaxInfo. Concentrating the selection inside query-relevant regions admits some visual similarity which is the cost of not spreading picks across irrelevant content.

\textit{\noindent\textbf{Takeaway.} No competing method simultaneously satisfies all three properties, while \textbf{FORGE} does, confirming that unifying relevance and diversity in a single geometry produces frame sets that no method optimizing a single property can match.}

\subsection{Efficiency}
\label{sec:exp-efficiency}

\begin{figure}[t]
    \centering
    \includegraphics[width=0.9\columnwidth]{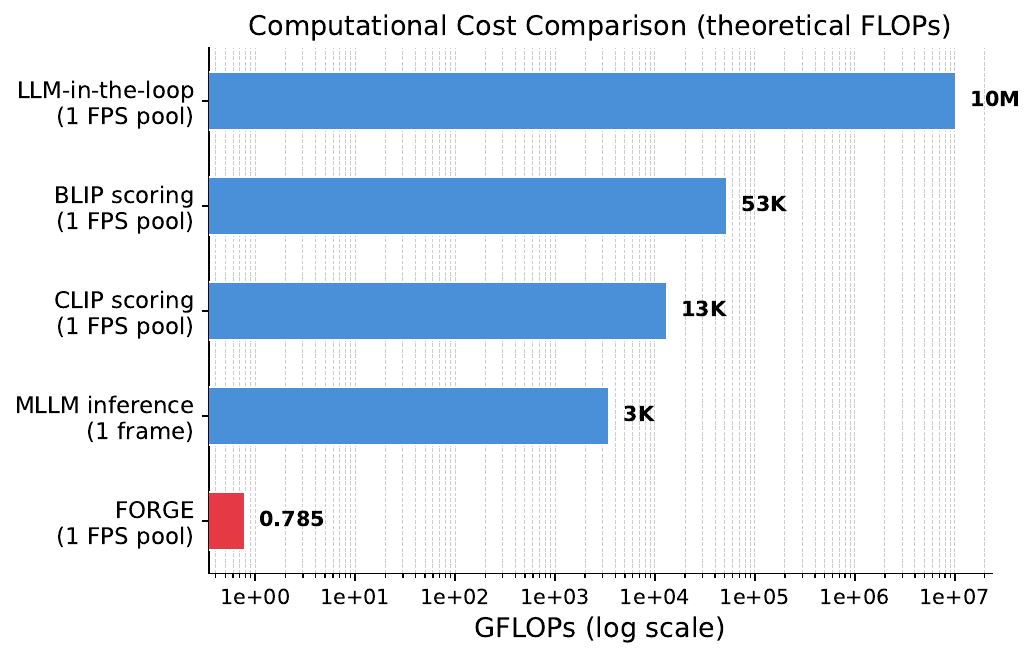}
    \caption{Computational cost (GFLOPs, log scale) of frame selection and inference. FLOPs are estimated on long videos (${\sim}$3,000--3,500 frames at 1 FPS, averaged over Video-MME and LongVideoBench). MLLM inference shows a single forward pass on one frame. \textbf{FORGE} operates entirely on pre-computed embeddings and scores, reducing selection-time compute by 6 orders of magnitude compared to BLIP-based scoring and over 7 orders of magnitude compared to LLM-in-the-loop approaches.}
    \label{fig:efficiency}
\end{figure}

\noindent\textbf{Takeaway.} \textbf{FORGE} achieves the accuracy improvements reported in the previous subsections at a fraction of the computational cost of LLM-in-the-loop methods, and at a cost comparable to uniform sampling.

\section{Conclusion}

\label{sec:conclusion}

Long-form video understanding with MLLMs is fundamentally an information problem. The evidence needed to answer a query is concentrated in a small fraction of the video duration, the remaining content is largely irrelevant to the query, and even within the relevant subsequence, frames are often redundant visually and semantically. In this work, we show that selecting frames with higher query-relevant information density leads to measurably better downstream accuracy across architectures, parameter scales, and benchmarks.

A multimodal embedding space encodes rich visual and semantic structure, but collapsing the query's high-dimensional structure into a scalar similarity score substitutes relevance with similarity and loses the structure needed to independently identify the frames that contain distinct query-relevant dimensions. \textbf{FORGE} resolves scalar collapse by inducing a query-conditioned geometry in which the query's effective dimensional structure is preserved, so that maximizing the information spanned in this geometry captures relevant and non-redundant content jointly. The ablation in Section~\ref{sec:exp-ablation} confirms that the geometry is the source of the performance gain.

\paragraph{Limitations.} Despite consistent improvements in downstream accuracy, a gap to reach a perfect video understanding remains. This gap derives from MLLMs' reasoning and cross-modality misalignment. Certain task categories, including counting, fine-grained object recognition, and complex spatial reasoning, remain challenging even when the model is presented with all the relevant content, as these reflect the model's training and architecture rather than its input. Selecting the right frames is a prerequisite for the reasoning these models still find difficult, not a substitute for it.

\backmatter

\section*{Declarations}

\bmhead{Availability of data and materials}
This study analyses the publicly available Video-MME~\citep{fu2025video} and
LongVideoBench~\citep{wu2024longvideobench} benchmarks, obtainable from their
official repositories. No new data were generated during this study. The
implementation will be released publicly upon acceptance.

\bmhead{Competing interests}
The authors declare no competing interests.

\bmhead{Funding}
This work is partially funded through the support from the Franklin Foundation
via the John and Marilu McCarty Chair professorship and a gift from Lab126 at
Amazon.

\bmhead{Use of generative artificial intelligence}
During the preparation of this manuscript, the authors used AI-assistant tools
for editing and revising the text grammar and tone and reviewing the content.
The authors reviewed and edited all suggested text and take full responsibility
for the content of this publication.




\begin{thebibliography}{66}
\ifx \bisbn   \undefined \def \bisbn  #1{ISBN #1}\fi
\ifx \binits  \undefined \def \binits#1{#1}\fi
\ifx \bauthor  \undefined \def \bauthor#1{#1}\fi
\ifx \batitle  \undefined \def \batitle#1{#1}\fi
\ifx \bjtitle  \undefined \def \bjtitle#1{#1}\fi
\ifx \bvolume  \undefined \def \bvolume#1{\textbf{#1}}\fi
\ifx \byear  \undefined \def \byear#1{#1}\fi
\ifx \bissue  \undefined \def \bissue#1{#1}\fi
\ifx \bfpage  \undefined \def \bfpage#1{#1}\fi
\ifx \blpage  \undefined \def \blpage #1{#1}\fi
\ifx \burl  \undefined \def \burl#1{\textsf{#1}}\fi
\ifx \doiurl  \undefined \def \doiurl#1{\url{https://doi.org/#1}}\fi
\ifx \betal  \undefined \def \betal{\textit{et al.}}\fi
\ifx \binstitute  \undefined \def \binstitute#1{#1}\fi
\ifx \binstitutionaled  \undefined \def \binstitutionaled#1{#1}\fi
\ifx \bctitle  \undefined \def \bctitle#1{#1}\fi
\ifx \beditor  \undefined \def \beditor#1{#1}\fi
\ifx \bpublisher  \undefined \def \bpublisher#1{#1}\fi
\ifx \bbtitle  \undefined \def \bbtitle#1{#1}\fi
\ifx \bedition  \undefined \def \bedition#1{#1}\fi
\ifx \bseriesno  \undefined \def \bseriesno#1{#1}\fi
\ifx \blocation  \undefined \def \blocation#1{#1}\fi
\ifx \bsertitle  \undefined \def \bsertitle#1{#1}\fi
\ifx \bsnm \undefined \def \bsnm#1{#1}\fi
\ifx \bsuffix \undefined \def \bsuffix#1{#1}\fi
\ifx \bparticle \undefined \def \bparticle#1{#1}\fi
\ifx \barticle \undefined \def \barticle#1{#1}\fi
\bibcommenthead
\ifx \bconfdate \undefined \def \bconfdate #1{#1}\fi
\ifx \botherref \undefined \def \botherref #1{#1}\fi
\ifx \url \undefined \def \url#1{\textsf{#1}}\fi
\ifx \bchapter \undefined \def \bchapter#1{#1}\fi
\ifx \bbook \undefined \def \bbook#1{#1}\fi
\ifx \bcomment \undefined \def \bcomment#1{#1}\fi
\ifx \oauthor \undefined \def \oauthor#1{#1}\fi
\ifx \citeauthoryear \undefined \def \citeauthoryear#1{#1}\fi
\ifx \endbibitem  \undefined \def \endbibitem {}\fi
\ifx \bconflocation  \undefined \def \bconflocation#1{#1}\fi
\ifx \arxivurl  \undefined \def \arxivurl#1{\textsf{#1}}\fi
\csname PreBibitemsHook\endcsname

\bibitem[\protect\citeauthoryear{Lin et~al.}{2024}]{lin2024video}
\begin{bchapter}
\bauthor{\bsnm{Lin}, \binits{B.}},
\bauthor{\bsnm{Ye}, \binits{Y.}},
\bauthor{\bsnm{Zhu}, \binits{B.}},
\bauthor{\bsnm{Cui}, \binits{J.}},
\bauthor{\bsnm{Ning}, \binits{M.}},
\bauthor{\bsnm{Jin}, \binits{P.}},
\bauthor{\bsnm{Yuan}, \binits{L.}}:
\bctitle{Video-llava: Learning united visual representation by alignment before
  projection}.
In: \bbtitle{Proceedings of the 2024 Conference on Empirical Methods in Natural
  Language Processing},
pp. \bfpage{5971}--\blpage{5984}
(\byear{2024})
\end{bchapter}
\endbibitem

\bibitem[\protect\citeauthoryear{Maaz et~al.}{2024}]{maaz2024video}
\begin{bchapter}
\bauthor{\bsnm{Maaz}, \binits{M.}},
\bauthor{\bsnm{Rasheed}, \binits{H.}},
\bauthor{\bsnm{Khan}, \binits{S.}},
\bauthor{\bsnm{Khan}, \binits{F.}}:
\bctitle{Video-chatgpt: Towards detailed video understanding via large vision
  and language models}.
In: \bbtitle{Proceedings of the 62nd Annual Meeting of the Association for
  Computational Linguistics (Volume 1: Long Papers)},
pp. \bfpage{12585}--\blpage{12602}
(\byear{2024})
\end{bchapter}
\endbibitem

\bibitem[\protect\citeauthoryear{Bai et~al.}{2025}]{bai2025qwen2}
\begin{botherref}
\oauthor{\bsnm{Bai}, \binits{S.}},
\oauthor{\bsnm{Chen}, \binits{K.}},
\oauthor{\bsnm{Liu}, \binits{X.}},
\oauthor{\bsnm{Wang}, \binits{J.}},
\oauthor{\bsnm{Ge}, \binits{W.}},
\oauthor{\bsnm{Song}, \binits{S.}},
\oauthor{\bsnm{Dang}, \binits{K.}},
\oauthor{\bsnm{Wang}, \binits{P.}},
\oauthor{\bsnm{Wang}, \binits{S.}},
\oauthor{\bsnm{Tang}, \binits{J.}}, et al.:
Qwen2. 5-vl technical report.
arXiv preprint arXiv:2502.13923
(2025)
\end{botherref}
\endbibitem

\bibitem[\protect\citeauthoryear{Chen et~al.}{2024}]{chen2024far}
\begin{barticle}
\bauthor{\bsnm{Chen}, \binits{Z.}},
\bauthor{\bsnm{Wang}, \binits{W.}},
\bauthor{\bsnm{Tian}, \binits{H.}},
\bauthor{\bsnm{Ye}, \binits{S.}},
\bauthor{\bsnm{Gao}, \binits{Z.}},
\bauthor{\bsnm{Cui}, \binits{E.}},
\bauthor{\bsnm{Tong}, \binits{W.}},
\bauthor{\bsnm{Hu}, \binits{K.}},
\bauthor{\bsnm{Luo}, \binits{J.}},
\bauthor{\bsnm{Ma}, \binits{Z.}}, \betal:
\batitle{How far are we to gpt-4v? closing the gap to commercial multimodal
  models with open-source suites}.
\bjtitle{Science China Information Sciences}
\bvolume{67}(\bissue{12}),
\bfpage{220101}
(\byear{2024})
\end{barticle}
\endbibitem

\bibitem[\protect\citeauthoryear{Fu et~al.}{2025}]{fu2025video}
\begin{bchapter}
\bauthor{\bsnm{Fu}, \binits{C.}},
\bauthor{\bsnm{Dai}, \binits{Y.}},
\bauthor{\bsnm{Luo}, \binits{Y.}},
\bauthor{\bsnm{Li}, \binits{L.}},
\bauthor{\bsnm{Ren}, \binits{S.}},
\bauthor{\bsnm{Zhang}, \binits{R.}},
\bauthor{\bsnm{Wang}, \binits{Z.}},
\bauthor{\bsnm{Zhou}, \binits{C.}},
\bauthor{\bsnm{Shen}, \binits{Y.}},
\bauthor{\bsnm{Zhang}, \binits{M.}}, \betal:
\bctitle{Video-mme: The first-ever comprehensive evaluation benchmark of
  multi-modal llms in video analysis}.
In: \bbtitle{Proceedings of the Computer Vision and Pattern Recognition
  Conference},
pp. \bfpage{24108}--\blpage{24118}
(\byear{2025})
\end{bchapter}
\endbibitem

\bibitem[\protect\citeauthoryear{Wu et~al.}{2024}]{wu2024longvideobench}
\begin{barticle}
\bauthor{\bsnm{Wu}, \binits{H.}},
\bauthor{\bsnm{Li}, \binits{D.}},
\bauthor{\bsnm{Chen}, \binits{B.}},
\bauthor{\bsnm{Li}, \binits{J.}}:
\batitle{Longvideobench: A benchmark for long-context interleaved
  video-language understanding}.
\bjtitle{Advances in Neural Information Processing Systems}
\bvolume{37},
\bfpage{28828}--\blpage{28857}
(\byear{2024})
\end{barticle}
\endbibitem

\bibitem[\protect\citeauthoryear{Li et~al.}{2026}]{li2026kfs}
\begin{bchapter}
\bauthor{\bsnm{Li}, \binits{Z.}},
\bauthor{\bsnm{Ishida}, \binits{K.}},
\bauthor{\bsnm{Yamazaki}, \binits{S.}},
\bauthor{\bsnm{Ji}, \binits{X.}},
\bauthor{\bsnm{Liu}, \binits{J.}}:
\bctitle{Kfs-bench: Comprehensive evaluation of key frame sampling in long
  video understanding}.
In: \bbtitle{Proceedings of the IEEE/CVF Winter Conference on Applications of
  Computer Vision},
pp. \bfpage{5643}--\blpage{5652}
(\byear{2026})
\end{bchapter}
\endbibitem

\bibitem[\protect\citeauthoryear{Shu et~al.}{2025}]{shu2025video}
\begin{bchapter}
\bauthor{\bsnm{Shu}, \binits{Y.}},
\bauthor{\bsnm{Liu}, \binits{Z.}},
\bauthor{\bsnm{Zhang}, \binits{P.}},
\bauthor{\bsnm{Qin}, \binits{M.}},
\bauthor{\bsnm{Zhou}, \binits{J.}},
\bauthor{\bsnm{Liang}, \binits{Z.}},
\bauthor{\bsnm{Huang}, \binits{T.}},
\bauthor{\bsnm{Zhao}, \binits{B.}}:
\bctitle{Video-xl: Extra-long vision language model for hour-scale video
  understanding}.
In: \bbtitle{Proceedings of the Computer Vision and Pattern Recognition
  Conference},
pp. \bfpage{26160}--\blpage{26169}
(\byear{2025})
\end{bchapter}
\endbibitem

\bibitem[\protect\citeauthoryear{Wang et~al.}{2024}]{wang2024videollamb}
\begin{botherref}
\oauthor{\bsnm{Wang}, \binits{Y.}},
\oauthor{\bsnm{Song}, \binits{Y.}},
\oauthor{\bsnm{Xie}, \binits{C.}},
\oauthor{\bsnm{Liu}, \binits{Y.}},
\oauthor{\bsnm{Zheng}, \binits{Z.}}:
Videollamb: Long streaming video understanding with recurrent memory bridges.
arXiv preprint arXiv:2409.01071
(2024)
\end{botherref}
\endbibitem

\bibitem[\protect\citeauthoryear{Jiang et~al.}{2025}]{jiang2025storm}
\begin{bchapter}
\bauthor{\bsnm{Jiang}, \binits{J.}},
\bauthor{\bsnm{Li}, \binits{X.}},
\bauthor{\bsnm{Liu}, \binits{Z.}},
\bauthor{\bsnm{Li}, \binits{M.}},
\bauthor{\bsnm{Chen}, \binits{G.}},
\bauthor{\bsnm{Li}, \binits{Z.}},
\bauthor{\bsnm{Huang}, \binits{D.-A.}},
\bauthor{\bsnm{Liu}, \binits{G.}},
\bauthor{\bsnm{Yu}, \binits{Z.}},
\bauthor{\bsnm{Keutzer}, \binits{K.}}, \betal:
\bctitle{Storm: Token-efficient long video understanding for multimodal llms}.
In: \bbtitle{Proceedings of the IEEE/CVF International Conference on Computer
  Vision},
pp. \bfpage{5830}--\blpage{5841}
(\byear{2025})
\end{bchapter}
\endbibitem

\bibitem[\protect\citeauthoryear{Liu et~al.}{2024}]{liu2024lost}
\begin{barticle}
\bauthor{\bsnm{Liu}, \binits{N.F.}},
\bauthor{\bsnm{Lin}, \binits{K.}},
\bauthor{\bsnm{Hewitt}, \binits{J.}},
\bauthor{\bsnm{Paranjape}, \binits{A.}},
\bauthor{\bsnm{Bevilacqua}, \binits{M.}},
\bauthor{\bsnm{Petroni}, \binits{F.}},
\bauthor{\bsnm{Liang}, \binits{P.}}:
\batitle{Lost in the middle: How language models use long contexts}.
\bjtitle{Transactions of the Association for Computational Linguistics}
\bvolume{12},
\bfpage{157}--\blpage{173}
(\byear{2024})
\end{barticle}
\endbibitem

\bibitem[\protect\citeauthoryear{Du et~al.}{2025}]{du-etal-2025-context}
\begin{bchapter}
\bauthor{\bsnm{Du}, \binits{Y.}},
\bauthor{\bsnm{Tian}, \binits{M.}},
\bauthor{\bsnm{Ronanki}, \binits{S.}},
\bauthor{\bsnm{Rongali}, \binits{S.}},
\bauthor{\bsnm{Bodapati}, \binits{S.B.}},
\bauthor{\bsnm{Galstyan}, \binits{A.}},
\bauthor{\bsnm{Wells}, \binits{A.}},
\bauthor{\bsnm{Schwartz}, \binits{R.}},
\bauthor{\bsnm{Huerta}, \binits{E.A.}},
\bauthor{\bsnm{Peng}, \binits{H.}}:
\bctitle{Context length alone hurts {LLM} performance despite perfect
  retrieval}.
In: \beditor{\bsnm{Christodoulopoulos}, \binits{C.}},
\beditor{\bsnm{Chakraborty}, \binits{T.}},
\beditor{\bsnm{Rose}, \binits{C.}},
\beditor{\bsnm{Peng}, \binits{V.}} (eds.)
\bbtitle{Findings of the Association for Computational Linguistics: EMNLP
  2025},
pp. \bfpage{23281}--\blpage{23298}.
\bpublisher{Association for Computational Linguistics},
\blocation{Suzhou, China}
(\byear{2025}).
\doiurl{10.18653/v1/2025.findings-emnlp.1264} .
\burl{https://aclanthology.org/2025.findings-emnlp.1264/}
\end{bchapter}
\endbibitem

\bibitem[\protect\citeauthoryear{Yang et~al.}{2025}]{Yang2025CVPR}
\begin{bchapter}
\bauthor{\bsnm{Yang}, \binits{S.}},
\bauthor{\bsnm{Chen}, \binits{Y.}},
\bauthor{\bsnm{Tian}, \binits{Z.}},
\bauthor{\bsnm{Wang}, \binits{C.}},
\bauthor{\bsnm{Li}, \binits{J.}},
\bauthor{\bsnm{Yu}, \binits{B.}},
\bauthor{\bsnm{Jia}, \binits{J.}}:
\bctitle{Visionzip: Longer is better but not necessary in vision language
  models}.
In: \bbtitle{Proceedings of the IEEE/CVF Conference on Computer Vision and
  Pattern Recognition (CVPR)},
pp. \bfpage{19792}--\blpage{19802}
(\byear{2025})
\end{bchapter}
\endbibitem

\bibitem[\protect\citeauthoryear{Shen et~al.}{2025}]{shen2025fastvid}
\begin{bchapter}
\bauthor{\bsnm{Shen}, \binits{L.}},
\bauthor{\bsnm{Gong}, \binits{G.}},
\bauthor{\bsnm{He}, \binits{T.}},
\bauthor{\bsnm{Zhang}, \binits{Y.}},
\bauthor{\bsnm{liu}},
\bauthor{\bsnm{Zhao}, \binits{S.}},
\bauthor{\bsnm{Ding}, \binits{G.}}:
\bctitle{Fast{VID}: Dynamic density pruning for fast video large language
  models}.
In: \bbtitle{The Thirty-ninth Annual Conference on Neural Information
  Processing Systems}
(\byear{2025}).
\burl{https://openreview.net/forum?id=2xS4VtpApy}
\end{bchapter}
\endbibitem

\bibitem[\protect\citeauthoryear{Wang et~al.}{2025}]{adaretake}
\begin{bchapter}
\bauthor{\bsnm{Wang}, \binits{X.}},
\bauthor{\bsnm{Si}, \binits{Q.}},
\bauthor{\bsnm{Zhu}, \binits{S.}},
\bauthor{\bsnm{Wu}, \binits{J.}},
\bauthor{\bsnm{Cao}, \binits{L.}},
\bauthor{\bsnm{Nie}, \binits{L.}}:
\bctitle{{A}da{R}e{T}a{K}e: Adaptive redundancy reduction to perceive longer
  for video-language understanding}.
In: \beditor{\bsnm{Che}, \binits{W.}},
\beditor{\bsnm{Nabende}, \binits{J.}},
\beditor{\bsnm{Shutova}, \binits{E.}},
\beditor{\bsnm{Pilehvar}, \binits{M.T.}} (eds.)
\bbtitle{Findings of the Association for Computational Linguistics: ACL 2025},
pp. \bfpage{5417}--\blpage{5432}.
\bpublisher{Association for Computational Linguistics},
\blocation{Vienna, Austria}
(\byear{2025}).
\doiurl{10.18653/v1/2025.findings-acl.283} .
\burl{https://aclanthology.org/2025.findings-acl.283/}
\end{bchapter}
\endbibitem

\bibitem[\protect\citeauthoryear{Wang et~al.}{2024}]{wang2024videoagent}
\begin{bchapter}
\bauthor{\bsnm{Wang}, \binits{X.}},
\bauthor{\bsnm{Zhang}, \binits{Y.}},
\bauthor{\bsnm{Zohar}, \binits{O.}},
\bauthor{\bsnm{Yeung-Levy}, \binits{S.}}:
\bctitle{Videoagent: Long-form video understanding with large language model as
  agent}.
In: \bbtitle{European Conference on Computer Vision},
pp. \bfpage{58}--\blpage{76}
(\byear{2024}).
\bcomment{Springer}
\end{bchapter}
\endbibitem

\bibitem[\protect\citeauthoryear{Zhang et~al.}{2025}]{zhang2025deep}
\begin{bchapter}
\bauthor{\bsnm{Zhang}, \binits{X.}},
\bauthor{\bsnm{Jia}, \binits{Z.}},
\bauthor{\bsnm{Guo}, \binits{Z.}},
\bauthor{\bsnm{Li}, \binits{J.}},
\bauthor{\bsnm{Li}, \binits{B.}},
\bauthor{\bsnm{Li}, \binits{H.}},
\bauthor{\bsnm{Lu}, \binits{Y.}}:
\bctitle{Deep video discovery: Agentic search with tool use for long-form video
  understanding}.
In: \bbtitle{The Thirty-ninth Annual Conference on Neural Information
  Processing Systems}
(\byear{2025}).
\burl{https://openreview.net/forum?id=oQYq9L1NVT}
\end{bchapter}
\endbibitem

\bibitem[\protect\citeauthoryear{Zou et~al.}{2025}]{zou2025air}
\begin{botherref}
\oauthor{\bsnm{Zou}, \binits{Y.}},
\oauthor{\bsnm{Jin}, \binits{S.}},
\oauthor{\bsnm{Deng}, \binits{A.}},
\oauthor{\bsnm{Zhao}, \binits{Y.}},
\oauthor{\bsnm{Wang}, \binits{J.}},
\oauthor{\bsnm{Chen}, \binits{C.}}:
Air: Enabling adaptive, iterative, and reasoning-based frame selection for
  video question answering.
arXiv preprint arXiv:2510.04428
(2025)
\end{botherref}
\endbibitem

\bibitem[\protect\citeauthoryear{Wang et~al.}{2025}]{wang2025avp}
\begin{botherref}
\oauthor{\bsnm{Wang}, \binits{Z.}},
\oauthor{\bsnm{Zhou}, \binits{H.}},
\oauthor{\bsnm{Wang}, \binits{S.}},
\oauthor{\bsnm{Li}, \binits{J.}},
\oauthor{\bsnm{Xiong}, \binits{C.}},
\oauthor{\bsnm{Savarese}, \binits{S.}},
\oauthor{\bsnm{Bansal}, \binits{M.}},
\oauthor{\bsnm{Ryoo}, \binits{M.S.}},
\oauthor{\bsnm{Niebles}, \binits{J.C.}}:
Active Video Perception: Iterative Evidence Seeking for Agentic Long Video
  Understanding
(2025).
\url{https://arxiv.org/abs/2512.05774}
\end{botherref}
\endbibitem

\bibitem[\protect\citeauthoryear{Tang et~al.}{2025}]{tang2025adaptive}
\begin{bchapter}
\bauthor{\bsnm{Tang}, \binits{X.}},
\bauthor{\bsnm{Qiu}, \binits{J.}},
\bauthor{\bsnm{Xie}, \binits{L.}},
\bauthor{\bsnm{Tian}, \binits{Y.}},
\bauthor{\bsnm{Jiao}, \binits{J.}},
\bauthor{\bsnm{Ye}, \binits{Q.}}:
\bctitle{Adaptive keyframe sampling for long video understanding}.
In: \bbtitle{Proceedings of the Computer Vision and Pattern Recognition
  Conference},
pp. \bfpage{29118}--\blpage{29128}
(\byear{2025})
\end{bchapter}
\endbibitem

\bibitem[\protect\citeauthoryear{Sun et~al.}{2025}]{Sun_2025_ICCV}
\begin{bchapter}
\bauthor{\bsnm{Sun}, \binits{H.}},
\bauthor{\bsnm{Lu}, \binits{S.}},
\bauthor{\bsnm{Wang}, \binits{H.}},
\bauthor{\bsnm{Chen}, \binits{Q.-G.}},
\bauthor{\bsnm{Xu}, \binits{Z.}},
\bauthor{\bsnm{Luo}, \binits{W.}},
\bauthor{\bsnm{Zhang}, \binits{K.}},
\bauthor{\bsnm{Li}, \binits{M.}}:
\bctitle{Mdp3: A training-free approach for list-wise frame selection in
  video-llms}.
In: \bbtitle{Proceedings of the IEEE/CVF International Conference on Computer
  Vision (ICCV)},
pp. \bfpage{24090}--\blpage{24101}
(\byear{2025})
\end{bchapter}
\endbibitem

\bibitem[\protect\citeauthoryear{Liu et~al.}{2025}]{liu2025bolt}
\begin{bchapter}
\bauthor{\bsnm{Liu}, \binits{S.}},
\bauthor{\bsnm{Zhao}, \binits{C.}},
\bauthor{\bsnm{Xu}, \binits{T.}},
\bauthor{\bsnm{Ghanem}, \binits{B.}}:
\bctitle{Bolt: Boost large vision-language model without training for long-form
  video understanding}.
In: \bbtitle{Proceedings of the Computer Vision and Pattern Recognition
  Conference},
pp. \bfpage{3318}--\blpage{3327}
(\byear{2025})
\end{bchapter}
\endbibitem

\bibitem[\protect\citeauthoryear{Li et~al.}{2025}]{li2025maxinfo}
\begin{botherref}
\oauthor{\bsnm{Li}, \binits{P.}},
\oauthor{\bsnm{Abdullaeva}, \binits{I.}},
\oauthor{\bsnm{Gambashidze}, \binits{A.}},
\oauthor{\bsnm{Kuznetsov}, \binits{A.}},
\oauthor{\bsnm{Oseledets}, \binits{I.}}:
Maxinfo: A training-free key-frame selection method using maximum volume for
  enhanced video understanding.
arXiv preprint arXiv:2502.03183
(2025)
\end{botherref}
\endbibitem

\bibitem[\protect\citeauthoryear{Zhang et~al.}{2025}]{Zhang_2025_ICCV}
\begin{bchapter}
\bauthor{\bsnm{Zhang}, \binits{S.}},
\bauthor{\bsnm{Yang}, \binits{J.}},
\bauthor{\bsnm{Yin}, \binits{J.}},
\bauthor{\bsnm{Luo}, \binits{Z.}},
\bauthor{\bsnm{Luan}, \binits{J.}}:
\bctitle{Q-frame: Query-aware frame selection and multi-resolution adaptation
  for video-llms}.
In: \bbtitle{Proceedings of the IEEE/CVF International Conference on Computer
  Vision (ICCV)},
pp. \bfpage{22056}--\blpage{22065}
(\byear{2025})
\end{bchapter}
\endbibitem

\bibitem[\protect\citeauthoryear{Weng et~al.}{2024}]{weng2024longvlm}
\begin{botherref}
\oauthor{\bsnm{Weng}, \binits{Y.}},
\oauthor{\bsnm{Han}, \binits{M.}},
\oauthor{\bsnm{He}, \binits{H.}},
\oauthor{\bsnm{Chang}, \binits{X.}},
\oauthor{\bsnm{Zhuang}, \binits{B.}}:
Longvlm: Efficient long video understanding via large language models.
European Conference on Computer Vision (ECCV)
(2024)
\end{botherref}
\endbibitem

\bibitem[\protect\citeauthoryear{Yu et~al.}{2023}]{yu2023self}
\begin{barticle}
\bauthor{\bsnm{Yu}, \binits{S.}},
\bauthor{\bsnm{Cho}, \binits{J.}},
\bauthor{\bsnm{Yadav}, \binits{P.}},
\bauthor{\bsnm{Bansal}, \binits{M.}}:
\batitle{Self-chained image-language model for video localization and question
  answering}.
\bjtitle{Advances in Neural Information Processing Systems}
\bvolume{36},
\bfpage{76749}--\blpage{76771}
(\byear{2023})
\end{barticle}
\endbibitem

\bibitem[\protect\citeauthoryear{Bai et~al.}{2025}]{bai2025qwen3vl}
\begin{botherref}
\oauthor{\bsnm{Bai}, \binits{S.}},
\oauthor{\bsnm{Cai}, \binits{Y.}},
\oauthor{\bsnm{Chen}, \binits{R.}},
\oauthor{\bsnm{Chen}, \binits{K.}},
\oauthor{\bsnm{Chen}, \binits{X.}},
\oauthor{\bsnm{Cheng}, \binits{Z.}},
\oauthor{\bsnm{Deng}, \binits{L.}},
\oauthor{\bsnm{Ding}, \binits{W.}},
\oauthor{\bsnm{Gao}, \binits{C.}},
\oauthor{\bsnm{Ge}, \binits{C.}},
\oauthor{\bsnm{Ge}, \binits{W.}},
\oauthor{\bsnm{Guo}, \binits{Z.}},
\oauthor{\bsnm{Huang}, \binits{Q.}},
\oauthor{\bsnm{Huang}, \binits{J.}},
\oauthor{\bsnm{Huang}, \binits{F.}},
\oauthor{\bsnm{Hui}, \binits{B.}},
\oauthor{\bsnm{Jiang}, \binits{S.}},
\oauthor{\bsnm{Li}, \binits{Z.}},
\oauthor{\bsnm{Li}, \binits{M.}},
\oauthor{\bsnm{Li}, \binits{M.}},
\oauthor{\bsnm{Li}, \binits{K.}},
\oauthor{\bsnm{Lin}, \binits{Z.}},
\oauthor{\bsnm{Lin}, \binits{J.}},
\oauthor{\bsnm{Liu}, \binits{X.}},
\oauthor{\bsnm{Liu}, \binits{J.}},
\oauthor{\bsnm{Liu}, \binits{C.}},
\oauthor{\bsnm{Liu}, \binits{Y.}},
\oauthor{\bsnm{Liu}, \binits{D.}},
\oauthor{\bsnm{Liu}, \binits{S.}},
\oauthor{\bsnm{Lu}, \binits{D.}},
\oauthor{\bsnm{Luo}, \binits{R.}},
\oauthor{\bsnm{Lv}, \binits{C.}},
\oauthor{\bsnm{Men}, \binits{R.}},
\oauthor{\bsnm{Meng}, \binits{L.}},
\oauthor{\bsnm{Ren}, \binits{X.}},
\oauthor{\bsnm{Ren}, \binits{X.}},
\oauthor{\bsnm{Song}, \binits{S.}},
\oauthor{\bsnm{Sun}, \binits{Y.}},
\oauthor{\bsnm{Tang}, \binits{J.}},
\oauthor{\bsnm{Tu}, \binits{J.}},
\oauthor{\bsnm{Wan}, \binits{J.}},
\oauthor{\bsnm{Wang}, \binits{P.}},
\oauthor{\bsnm{Wang}, \binits{Q.}},
\oauthor{\bsnm{Wang}, \binits{Y.}},
\oauthor{\bsnm{Xie}, \binits{T.}},
\oauthor{\bsnm{Xu}, \binits{Y.}},
\oauthor{\bsnm{Xu}, \binits{H.}},
\oauthor{\bsnm{Xu}, \binits{J.}},
\oauthor{\bsnm{Yang}, \binits{Z.}},
\oauthor{\bsnm{Yang}, \binits{M.}},
\oauthor{\bsnm{Yang}, \binits{J.}},
\oauthor{\bsnm{Yang}, \binits{A.}},
\oauthor{\bsnm{Yu}, \binits{B.}},
\oauthor{\bsnm{Zhang}, \binits{F.}},
\oauthor{\bsnm{Zhang}, \binits{H.}},
\oauthor{\bsnm{Zhang}, \binits{X.}},
\oauthor{\bsnm{Zheng}, \binits{B.}},
\oauthor{\bsnm{Zhong}, \binits{H.}},
\oauthor{\bsnm{Zhou}, \binits{J.}},
\oauthor{\bsnm{Zhou}, \binits{F.}},
\oauthor{\bsnm{Zhou}, \binits{J.}}:
Qwen3-vl technical report.
arXiv preprint arXiv:2511.21631
(2025)
\end{botherref}
\endbibitem

\bibitem[\protect\citeauthoryear{Yang et~al.}{2025}]{Yang_2025_ICCV}
\begin{bchapter}
\bauthor{\bsnm{Yang}, \binits{S.}},
\bauthor{\bsnm{Xu}, \binits{R.}},
\bauthor{\bsnm{Cui}, \binits{C.}},
\bauthor{\bsnm{Wang}, \binits{T.}},
\bauthor{\bsnm{Lin}, \binits{D.}},
\bauthor{\bsnm{Pang}, \binits{J.}}:
\bctitle{Vflowopt: A token pruning framework for lmms with visual information
  flow-guided optimization}.
In: \bbtitle{Proceedings of the IEEE/CVF International Conference on Computer
  Vision (ICCV)},
pp. \bfpage{23924}--\blpage{23934}
(\byear{2025})
\end{bchapter}
\endbibitem

\bibitem[\protect\citeauthoryear{Huang et~al.}{2025}]{prunevid}
\begin{bchapter}
\bauthor{\bsnm{Huang}, \binits{X.}},
\bauthor{\bsnm{Zhou}, \binits{H.}},
\bauthor{\bsnm{Han}, \binits{K.}}:
\bctitle{{P}rune{V}id: Visual token pruning for efficient video large language
  models}.
In: \beditor{\bsnm{Che}, \binits{W.}},
\beditor{\bsnm{Nabende}, \binits{J.}},
\beditor{\bsnm{Shutova}, \binits{E.}},
\beditor{\bsnm{Pilehvar}, \binits{M.T.}} (eds.)
\bbtitle{Findings of the Association for Computational Linguistics: ACL 2025},
pp. \bfpage{19959}--\blpage{19973}.
\bpublisher{Association for Computational Linguistics},
\blocation{Vienna, Austria}
(\byear{2025}).
\doiurl{10.18653/v1/2025.findings-acl.1024} .
\burl{https://aclanthology.org/2025.findings-acl.1024/}
\end{bchapter}
\endbibitem

\bibitem[\protect\citeauthoryear{Zhang et~al.}{2025}]{zhang2025sparsevlm}
\begin{bchapter}
\bauthor{\bsnm{Zhang}, \binits{Y.}},
\bauthor{\bsnm{Fan}, \binits{C.-K.}},
\bauthor{\bsnm{Ma}, \binits{J.}},
\bauthor{\bsnm{Zheng}, \binits{W.}},
\bauthor{\bsnm{Huang}, \binits{T.}},
\bauthor{\bsnm{Cheng}, \binits{K.}},
\bauthor{\bsnm{Gudovskiy}, \binits{D.A.}},
\bauthor{\bsnm{Okuno}, \binits{T.}},
\bauthor{\bsnm{Nakata}, \binits{Y.}},
\bauthor{\bsnm{Keutzer}, \binits{K.}},
\bauthor{\bsnm{Zhang}, \binits{S.}}:
\bctitle{Sparse{VLM}: Visual token sparsification for efficient vision-language
  model inference}.
In: \bbtitle{Forty-second International Conference on Machine Learning}
(\byear{2025}).
\burl{https://openreview.net/forum?id=80faIPZ67S}
\end{bchapter}
\endbibitem

\bibitem[\protect\citeauthoryear{Tong et~al.}{2025}]{tong2025flowcut}
\begin{bchapter}
\bauthor{\bsnm{Tong}, \binits{J.}},
\bauthor{\bsnm{Jin}, \binits{W.}},
\bauthor{\bsnm{Qin}, \binits{P.}},
\bauthor{\bsnm{Li}, \binits{A.}},
\bauthor{\bsnm{Zou}, \binits{Y.}},
\bauthor{\bsnm{Li}, \binits{Y.}},
\bauthor{\bsnm{Li}, \binits{Y.}},
\bauthor{\bsnm{Li}, \binits{R.}}:
\bctitle{Flowcut: Rethinking redundancy via information flow for efficient
  vision-language models}.
In: \bbtitle{The Thirty-ninth Annual Conference on Neural Information
  Processing Systems}
(\byear{2025}).
\burl{https://openreview.net/forum?id=M6zQNbCaLl}
\end{bchapter}
\endbibitem

\bibitem[\protect\citeauthoryear{Rao et~al.}{2021}]{rao2021dynamicvit}
\begin{barticle}
\bauthor{\bsnm{Rao}, \binits{Y.}},
\bauthor{\bsnm{Zhao}, \binits{W.}},
\bauthor{\bsnm{Liu}, \binits{B.}},
\bauthor{\bsnm{Lu}, \binits{J.}},
\bauthor{\bsnm{Zhou}, \binits{J.}},
\bauthor{\bsnm{Hsieh}, \binits{C.-J.}}:
\batitle{Dynamicvit: Efficient vision transformers with dynamic token
  sparsification}.
\bjtitle{Advances in neural information processing systems}
\bvolume{34},
\bfpage{13937}--\blpage{13949}
(\byear{2021})
\end{barticle}
\endbibitem

\bibitem[\protect\citeauthoryear{Wang et~al.}{2022}]{wang2022efficient}
\begin{bchapter}
\bauthor{\bsnm{Wang}, \binits{J.}},
\bauthor{\bsnm{Yang}, \binits{X.}},
\bauthor{\bsnm{Li}, \binits{H.}},
\bauthor{\bsnm{Liu}, \binits{L.}},
\bauthor{\bsnm{Wu}, \binits{Z.}},
\bauthor{\bsnm{Jiang}, \binits{Y.-G.}}:
\bctitle{Efficient video transformers with spatial-temporal token selection}.
In: \bbtitle{European Conference on Computer Vision},
pp. \bfpage{69}--\blpage{86}
(\byear{2022}).
\bcomment{Springer}
\end{bchapter}
\endbibitem

\bibitem[\protect\citeauthoryear{Fan et~al.}{2024}]{fan2024videoagent}
\begin{bchapter}
\bauthor{\bsnm{Fan}, \binits{Y.}},
\bauthor{\bsnm{Ma}, \binits{X.}},
\bauthor{\bsnm{Wu}, \binits{R.}},
\bauthor{\bsnm{Du}, \binits{Y.}},
\bauthor{\bsnm{Li}, \binits{J.}},
\bauthor{\bsnm{Gao}, \binits{Z.}},
\bauthor{\bsnm{Li}, \binits{Q.}}:
\bctitle{Videoagent: A memory-augmented multimodal agent for video
  understanding}.
In: \bbtitle{European Conference on Computer Vision},
pp. \bfpage{75}--\blpage{92}
(\byear{2024}).
\bcomment{Springer}
\end{bchapter}
\endbibitem

\bibitem[\protect\citeauthoryear{Yu et~al.}{2024}]{yu2024frame}
\begin{botherref}
\oauthor{\bsnm{Yu}, \binits{S.}},
\oauthor{\bsnm{Jin}, \binits{C.}},
\oauthor{\bsnm{Wang}, \binits{H.}},
\oauthor{\bsnm{Chen}, \binits{Z.}},
\oauthor{\bsnm{Jin}, \binits{S.}},
\oauthor{\bsnm{Zuo}, \binits{Z.}},
\oauthor{\bsnm{Xu}, \binits{X.}},
\oauthor{\bsnm{Sun}, \binits{Z.}},
\oauthor{\bsnm{Zhang}, \binits{B.}},
\oauthor{\bsnm{Wu}, \binits{J.}}, et al.:
Frame-voyager: Learning to query frames for video large language models.
arXiv preprint arXiv:2410.03226
(2024)
\end{botherref}
\endbibitem

\bibitem[\protect\citeauthoryear{Ye et~al.}{2025}]{ye2025re}
\begin{bchapter}
\bauthor{\bsnm{Ye}, \binits{J.}},
\bauthor{\bsnm{Wang}, \binits{Z.}},
\bauthor{\bsnm{Sun}, \binits{H.}},
\bauthor{\bsnm{Chandrasegaran}, \binits{K.}},
\bauthor{\bsnm{Durante}, \binits{Z.}},
\bauthor{\bsnm{Eyzaguirre}, \binits{C.}},
\bauthor{\bsnm{Bisk}, \binits{Y.}},
\bauthor{\bsnm{Niebles}, \binits{J.C.}},
\bauthor{\bsnm{Adeli}, \binits{E.}},
\bauthor{\bsnm{Fei-Fei}, \binits{L.}}, \betal:
\bctitle{Re-thinking temporal search for long-form video understanding}.
In: \bbtitle{Proceedings of the Computer Vision and Pattern Recognition
  Conference},
pp. \bfpage{8579}--\blpage{8591}
(\byear{2025})
\end{bchapter}
\endbibitem

\bibitem[\protect\citeauthoryear{Wang et~al.}{2025}]{wang2025videotree}
\begin{bchapter}
\bauthor{\bsnm{Wang}, \binits{Z.}},
\bauthor{\bsnm{Yu}, \binits{S.}},
\bauthor{\bsnm{Stengel-Eskin}, \binits{E.}},
\bauthor{\bsnm{Yoon}, \binits{J.}},
\bauthor{\bsnm{Cheng}, \binits{F.}},
\bauthor{\bsnm{Bertasius}, \binits{G.}},
\bauthor{\bsnm{Bansal}, \binits{M.}}:
\bctitle{Videotree: Adaptive tree-based video representation for llm reasoning
  on long videos}.
In: \bbtitle{Proceedings of the Computer Vision and Pattern Recognition
  Conference},
pp. \bfpage{3272}--\blpage{3283}
(\byear{2025})
\end{bchapter}
\endbibitem

\bibitem[\protect\citeauthoryear{Wu et~al.}{2019a}]{wu2019adaframe}
\begin{botherref}
\oauthor{\bsnm{Wu}, \binits{Z.}},
\oauthor{\bsnm{Xiong}, \binits{C.}},
\oauthor{\bsnm{Ma}, \binits{C.-Y.}},
\oauthor{\bsnm{Socher}, \binits{R.}},
\oauthor{\bsnm{Davis}, \binits{L.S.}}:
AdaFrame: Adaptive Frame Selection for Fast Video Recognition
(2019).
\url{https://arxiv.org/abs/1811.12432}
\end{botherref}
\endbibitem

\bibitem[\protect\citeauthoryear{Wu et~al.}{2019b}]{wu2019MARL}
\begin{botherref}
\oauthor{\bsnm{Wu}, \binits{W.}},
\oauthor{\bsnm{He}, \binits{D.}},
\oauthor{\bsnm{Tan}, \binits{X.}},
\oauthor{\bsnm{Chen}, \binits{S.}},
\oauthor{\bsnm{Wen}, \binits{S.}}:
Multi-Agent Reinforcement Learning Based Frame Sampling for Effective Untrimmed
  Video Recognition
(2019).
\url{https://arxiv.org/abs/1907.13369}
\end{botherref}
\endbibitem

\bibitem[\protect\citeauthoryear{Lei et~al.}{2021}]{lei2021less}
\begin{bchapter}
\bauthor{\bsnm{Lei}, \binits{J.}},
\bauthor{\bsnm{Li}, \binits{L.}},
\bauthor{\bsnm{Zhou}, \binits{L.}},
\bauthor{\bsnm{Gan}, \binits{Z.}},
\bauthor{\bsnm{Berg}, \binits{T.L.}},
\bauthor{\bsnm{Bansal}, \binits{M.}},
\bauthor{\bsnm{Liu}, \binits{J.}}:
\bctitle{Less is more: Clipbert for video-and-language learning via sparse
  sampling}.
In: \bbtitle{Proceedings of the IEEE/CVF Conference on Computer Vision and
  Pattern Recognition},
pp. \bfpage{7331}--\blpage{7341}
(\byear{2021})
\end{bchapter}
\endbibitem

\bibitem[\protect\citeauthoryear{Buch et~al.}{2025}]{buch2025flexible}
\begin{bchapter}
\bauthor{\bsnm{Buch}, \binits{S.}},
\bauthor{\bsnm{Nagrani}, \binits{A.}},
\bauthor{\bsnm{Arnab}, \binits{A.}},
\bauthor{\bsnm{Schmid}, \binits{C.}}:
\bctitle{Flexible frame selection for efficient video reasoning}.
In: \bbtitle{Proceedings of the Computer Vision and Pattern Recognition
  Conference},
pp. \bfpage{29071}--\blpage{29082}
(\byear{2025})
\end{bchapter}
\endbibitem

\bibitem[\protect\citeauthoryear{Hu et~al.}{2025}]{Hu_2025_CVPR}
\begin{bchapter}
\bauthor{\bsnm{Hu}, \binits{K.}},
\bauthor{\bsnm{Gao}, \binits{F.}},
\bauthor{\bsnm{Nie}, \binits{X.}},
\bauthor{\bsnm{Zhou}, \binits{P.}},
\bauthor{\bsnm{Tran}, \binits{S.}},
\bauthor{\bsnm{Neiman}, \binits{T.}},
\bauthor{\bsnm{Wang}, \binits{L.}},
\bauthor{\bsnm{Shah}, \binits{M.}},
\bauthor{\bsnm{Hamid}, \binits{R.}},
\bauthor{\bsnm{Yin}, \binits{B.}},
\bauthor{\bsnm{Chilimbi}, \binits{T.}}:
\bctitle{M-llm based video frame selection for efficient video understanding}.
In: \bbtitle{Proceedings of the IEEE/CVF Conference on Computer Vision and
  Pattern Recognition (CVPR)},
pp. \bfpage{13702}--\blpage{13712}
(\byear{2025})
\end{bchapter}
\endbibitem

\bibitem[\protect\citeauthoryear{Son et~al.}{2024}]{son2024csta}
\begin{bchapter}
\bauthor{\bsnm{Son}, \binits{J.}},
\bauthor{\bsnm{Park}, \binits{J.}},
\bauthor{\bsnm{Kim}, \binits{K.}}:
\bctitle{Csta: Cnn-based spatiotemporal attention for video summarization}.
In: \bbtitle{Proceedings of the IEEE/CVF Conference on Computer Vision and
  Pattern Recognition},
pp. \bfpage{18847}--\blpage{18856}
(\byear{2024})
\end{bchapter}
\endbibitem

\bibitem[\protect\citeauthoryear{Hsu et~al.}{2023}]{10124837}
\begin{barticle}
\bauthor{\bsnm{Hsu}, \binits{T.-C.}},
\bauthor{\bsnm{Liao}, \binits{Y.-S.}},
\bauthor{\bsnm{Huang}, \binits{C.-R.}}:
\batitle{Video summarization with spatiotemporal vision transformer}.
\bjtitle{IEEE Transactions on Image Processing}
\bvolume{32},
\bfpage{3013}--\blpage{3026}
(\byear{2023})
\doiurl{10.1109/TIP.2023.3275069}
\end{barticle}
\endbibitem

\bibitem[\protect\citeauthoryear{Potapov
  et~al.}{2014}]{10.1007/978-3-319-10599-4_35}
\begin{bchapter}
\bauthor{\bsnm{Potapov}, \binits{D.}},
\bauthor{\bsnm{Douze}, \binits{M.}},
\bauthor{\bsnm{Harchaoui}, \binits{Z.}},
\bauthor{\bsnm{Schmid}, \binits{C.}}:
\bctitle{Category-specific video summarization}.
In: \beditor{\bsnm{Fleet}, \binits{D.}},
\beditor{\bsnm{Pajdla}, \binits{T.}},
\beditor{\bsnm{Schiele}, \binits{B.}},
\beditor{\bsnm{Tuytelaars}, \binits{T.}} (eds.)
\bbtitle{Computer Vision -- ECCV 2014},
pp. \bfpage{540}--\blpage{555}.
\bpublisher{Springer},
\blocation{Cham}
(\byear{2014})
\end{bchapter}
\endbibitem

\bibitem[\protect\citeauthoryear{Gygli et~al.}{2015}]{gygli2015video}
\begin{bchapter}
\bauthor{\bsnm{Gygli}, \binits{M.}},
\bauthor{\bsnm{Grabner}, \binits{H.}},
\bauthor{\bsnm{Van~Gool}, \binits{L.}}:
\bctitle{Video summarization by learning submodular mixtures of objectives}.
In: \bbtitle{Proceedings of the IEEE Conference on Computer Vision and Pattern
  Recognition},
pp. \bfpage{3090}--\blpage{3098}
(\byear{2015})
\end{bchapter}
\endbibitem

\bibitem[\protect\citeauthoryear{Gong et~al.}{2014}]{gong2014diverse}
\begin{botherref}
\oauthor{\bsnm{Gong}, \binits{B.}},
\oauthor{\bsnm{Chao}, \binits{W.-L.}},
\oauthor{\bsnm{Grauman}, \binits{K.}},
\oauthor{\bsnm{Sha}, \binits{F.}}:
Diverse sequential subset selection for supervised video summarization.
Advances in neural information processing systems
\textbf{27}
(2014)
\end{botherref}
\endbibitem

\bibitem[\protect\citeauthoryear{Sharghi et~al.}{2017}]{sharghi2017query}
\begin{bchapter}
\bauthor{\bsnm{Sharghi}, \binits{A.}},
\bauthor{\bsnm{Laurel}, \binits{J.S.}},
\bauthor{\bsnm{Gong}, \binits{B.}}:
\bctitle{Query-focused video summarization: Dataset, evaluation, and a memory
  network based approach}.
In: \bbtitle{Proceedings of the IEEE Conference on Computer Vision and Pattern
  Recognition},
pp. \bfpage{4788}--\blpage{4797}
(\byear{2017})
\end{bchapter}
\endbibitem

\bibitem[\protect\citeauthoryear{Huang and Worring}{2020}]{huang2020query}
\begin{bchapter}
\bauthor{\bsnm{Huang}, \binits{J.-H.}},
\bauthor{\bsnm{Worring}, \binits{M.}}:
\bctitle{Query-controllable video summarization}.
In: \bbtitle{Proceedings of the 2020 International Conference on Multimedia
  Retrieval},
pp. \bfpage{242}--\blpage{250}
(\byear{2020})
\end{bchapter}
\endbibitem

\bibitem[\protect\citeauthoryear{Huang et~al.}{2023}]{huang2023query}
\begin{bchapter}
\bauthor{\bsnm{Huang}, \binits{J.-H.}},
\bauthor{\bsnm{Murn}, \binits{L.}},
\bauthor{\bsnm{Mrak}, \binits{M.}},
\bauthor{\bsnm{Worring}, \binits{M.}}:
\bctitle{Query-based video summarization with pseudo label supervision}.
In: \bbtitle{2023 IEEE International Conference on Image Processing (ICIP)},
pp. \bfpage{1430}--\blpage{1434}
(\byear{2023}).
\bcomment{IEEE}
\end{bchapter}
\endbibitem

\bibitem[\protect\citeauthoryear{Zheng et~al.}{2026}]{zheng2026tifre}
\begin{botherref}
\oauthor{\bsnm{Zheng}, \binits{X.}},
\oauthor{\bsnm{Wang}, \binits{Z.}},
\oauthor{\bsnm{Peng}, \binits{Y.}}:
Tifre: Text-guided video frame reduction for efficient video multi-modal large
  language models.
arXiv preprint arXiv:2602.08861
(2026)
\end{botherref}
\endbibitem

\bibitem[\protect\citeauthoryear{Li
  et~al.}{2025}]{li2025dividegroundadaptingframe}
\begin{botherref}
\oauthor{\bsnm{Li}, \binits{J.}},
\oauthor{\bsnm{Li}, \binits{B.}},
\oauthor{\bsnm{Li}, \binits{J.}},
\oauthor{\bsnm{Lu}, \binits{Y.}}:
Divide, then Ground: Adapting Frame Selection to Query Types for Long-Form
  Video Understanding
(2025).
\url{https://arxiv.org/abs/2512.04000}
\end{botherref}
\endbibitem

\bibitem[\protect\citeauthoryear{Tan et~al.}{2026}]{tan2026think}
\begin{botherref}
\oauthor{\bsnm{Tan}, \binits{W.}},
\oauthor{\bsnm{Song}, \binits{R.}},
\oauthor{\bsnm{Li}, \binits{J.}},
\oauthor{\bsnm{Ju}, \binits{J.}},
\oauthor{\bsnm{Luo}, \binits{Z.}}:
Think-clip-sample: Slow-fast frame selection for video understanding.
arXiv preprint arXiv:2601.11359
(2026)
\end{botherref}
\endbibitem

\bibitem[\protect\citeauthoryear{Song et~al.}{2026}]{song2026ktv}
\begin{botherref}
\oauthor{\bsnm{Song}, \binits{B.}},
\oauthor{\bsnm{Peng}, \binits{J.}},
\oauthor{\bsnm{Zhang}, \binits{Y.}},
\oauthor{\bsnm{Chen}, \binits{G.}},
\oauthor{\bsnm{Yang}, \binits{F.}},
\oauthor{\bsnm{Guo}, \binits{J.}}:
Ktv: Keyframes and key tokens selection for efficient training-free video llms.
arXiv preprint arXiv:2602.03615
(2026)
\end{botherref}
\endbibitem

\bibitem[\protect\citeauthoryear{Sheng et~al.}{2026}]{sheng2026sevices}
\begin{botherref}
\oauthor{\bsnm{Sheng}, \binits{Y.}},
\oauthor{\bsnm{Hao}, \binits{Y.}},
\oauthor{\bsnm{Li}, \binits{C.}},
\oauthor{\bsnm{Wang}, \binits{S.}},
\oauthor{\bsnm{He}, \binits{X.}}:
SeVi{CES}: Unifying Semantic-Visual Evidence Consensus for Long Video
  Understanding
(2026).
\url{https://openreview.net/forum?id=L0dORRnUhu}
\end{botherref}
\endbibitem

\bibitem[\protect\citeauthoryear{Kulesza
  et~al.}{2012}]{kulesza2012determinantal}
\begin{barticle}
\bauthor{\bsnm{Kulesza}, \binits{A.}},
\bauthor{\bsnm{Taskar}, \binits{B.}}, \betal:
\batitle{Determinantal point processes for machine learning}.
\bjtitle{Foundations and Trends{\textregistered} in Machine Learning}
\bvolume{5}(\bissue{2--3}),
\bfpage{123}--\blpage{286}
(\byear{2012})
\end{barticle}
\endbibitem

\bibitem[\protect\citeauthoryear{Goreinov et~al.}{2010}]{goreinov2010find}
\begin{bchapter}
\bauthor{\bsnm{Goreinov}, \binits{S.A.}},
\bauthor{\bsnm{Oseledets}, \binits{I.V.}},
\bauthor{\bsnm{Savostyanov}, \binits{D.V.}},
\bauthor{\bsnm{Tyrtyshnikov}, \binits{E.E.}},
\bauthor{\bsnm{Zamarashkin}, \binits{N.L.}}:
\bctitle{How to find a good submatrix}.
In: \bbtitle{Matrix Methods: Theory, Algorithms and Applications},
pp. \bfpage{247}--\blpage{256}.
\bpublisher{World Scientific}, \blocation{???}
(\byear{2010})
\end{bchapter}
\endbibitem

\bibitem[\protect\citeauthoryear{Mikhalev and
  Oseledets}{2018}]{mikhalev2018rectangular}
\begin{barticle}
\bauthor{\bsnm{Mikhalev}, \binits{A.}},
\bauthor{\bsnm{Oseledets}, \binits{I.V.}}:
\batitle{Rectangular maximum-volume submatrices and their applications}.
\bjtitle{Linear Algebra and its Applications}
\bvolume{538},
\bfpage{187}--\blpage{211}
(\byear{2018})
\end{barticle}
\endbibitem

\bibitem[\protect\citeauthoryear{Nemhauser
  et~al.}{1978}]{nemhauser1978analysis}
\begin{barticle}
\bauthor{\bsnm{Nemhauser}, \binits{G.L.}},
\bauthor{\bsnm{Wolsey}, \binits{L.A.}},
\bauthor{\bsnm{Fisher}, \binits{M.L.}}:
\batitle{An analysis of approximations for maximizing submodular set
  functions—i}.
\bjtitle{Mathematical programming}
\bvolume{14}(\bissue{1}),
\bfpage{265}--\blpage{294}
(\byear{1978})
\end{barticle}
\endbibitem

\bibitem[\protect\citeauthoryear{Cover and Thomas}{2006}]{cover2006elements}
\begin{bbook}
\bauthor{\bsnm{Cover}, \binits{T.M.}},
\bauthor{\bsnm{Thomas}, \binits{J.A.}}:
\bbtitle{Elements of Information Theory},
\bedition{2}nd edn.
\bpublisher{Wiley-Interscience}, \blocation{???}
(\byear{2006})
\end{bbook}
\endbibitem

\bibitem[\protect\citeauthoryear{Lindeberg}{1998}]{lindeberg1998feature}
\begin{barticle}
\bauthor{\bsnm{Lindeberg}, \binits{T.}}:
\batitle{Feature detection with automatic scale selection}.
\bjtitle{International Journal of Computer Vision}
\bvolume{30}(\bissue{2}),
\bfpage{79}--\blpage{116}
(\byear{1998})
\end{barticle}
\endbibitem

\bibitem[\protect\citeauthoryear{{Gemma Team}}{2025}]{gemma32025}
\begin{botherref}
\oauthor{\bsnm{{Gemma Team}}}:
Gemma 3 technical report.
arXiv preprint arXiv:2503.19786
(2025)
\end{botherref}
\endbibitem

\bibitem[\protect\citeauthoryear{Li et~al.}{2024}]{li2024llava}
\begin{botherref}
\oauthor{\bsnm{Li}, \binits{B.}},
\oauthor{\bsnm{Zhang}, \binits{Y.}},
\oauthor{\bsnm{Guo}, \binits{D.}},
\oauthor{\bsnm{Zhang}, \binits{R.}},
\oauthor{\bsnm{Li}, \binits{F.}},
\oauthor{\bsnm{Zhang}, \binits{H.}},
\oauthor{\bsnm{Zhang}, \binits{K.}},
\oauthor{\bsnm{Li}, \binits{Y.}},
\oauthor{\bsnm{Liu}, \binits{Z.}},
\oauthor{\bsnm{Li}, \binits{C.}}:
Llava-onevision: Easy visual task transfer.
arXiv preprint arXiv:2408.03326
(2024)
\end{botherref}
\endbibitem

\bibitem[\protect\citeauthoryear{Zhang et~al.}{2024}]{zhang2024long}
\begin{botherref}
\oauthor{\bsnm{Zhang}, \binits{P.}},
\oauthor{\bsnm{Zhang}, \binits{K.}},
\oauthor{\bsnm{Li}, \binits{B.}},
\oauthor{\bsnm{Zeng}, \binits{G.}},
\oauthor{\bsnm{Yang}, \binits{J.}},
\oauthor{\bsnm{Zhang}, \binits{Y.}},
\oauthor{\bsnm{Wang}, \binits{Z.}},
\oauthor{\bsnm{Tan}, \binits{H.}},
\oauthor{\bsnm{Li}, \binits{C.}},
\oauthor{\bsnm{Liu}, \binits{Z.}}:
Long context transfer from language to vision.
arXiv preprint arXiv:2406.16852
(2024)
\end{botherref}
\endbibitem

\bibitem[\protect\citeauthoryear{Wang et~al.}{2025}]{wang2025internvideo25}
\begin{botherref}
\oauthor{\bsnm{Wang}, \binits{Y.}},
\oauthor{\bsnm{Li}, \binits{X.}},
\oauthor{\bsnm{Yan}, \binits{Z.}},
\oauthor{\bsnm{He}, \binits{Y.}},
\oauthor{\bsnm{Yu}, \binits{J.}},
\oauthor{\bsnm{Zeng}, \binits{X.}},
\oauthor{\bsnm{Wang}, \binits{C.}},
\oauthor{\bsnm{Ma}, \binits{C.}},
\oauthor{\bsnm{Huang}, \binits{H.}},
\oauthor{\bsnm{Gao}, \binits{J.}},
\oauthor{\bsnm{Dou}, \binits{M.}},
\oauthor{\bsnm{Chen}, \binits{K.}},
\oauthor{\bsnm{Wang}, \binits{W.}},
\oauthor{\bsnm{Qiao}, \binits{Y.}},
\oauthor{\bsnm{Wang}, \binits{Y.}},
\oauthor{\bsnm{Wang}, \binits{L.}}:
Internvideo2.5: Empowering video mllms with long and rich context modeling.
arXiv preprint arXiv:2501.12386
(2025)
\end{botherref}
\endbibitem

\bibitem[\protect\citeauthoryear{Chen et~al.}{2024}]{chen2024internvl25}
\begin{botherref}
\oauthor{\bsnm{Chen}, \binits{Z.}},
\oauthor{\bsnm{Wang}, \binits{W.}},
\oauthor{\bsnm{Cao}, \binits{Y.}},
\oauthor{\bsnm{Liu}, \binits{Y.}},
\oauthor{\bsnm{Gao}, \binits{Z.}},
\oauthor{\bsnm{Cui}, \binits{E.}},
\oauthor{\bsnm{Zhu}, \binits{J.}},
\oauthor{\bsnm{Ye}, \binits{S.}},
\oauthor{\bsnm{Tian}, \binits{H.}},
\oauthor{\bsnm{Liu}, \binits{Z.}}, et al.:
Expanding performance boundaries of open-source multimodal models with model,
  data, and test-time scaling.
arXiv preprint arXiv:2412.05271
(2024)
\end{botherref}
\endbibitem

\end{thebibliography}
\end{document}